\documentclass{article}

\usepackage[final]{neurips_2024}

\usepackage[utf8]{inputenc} % allow utf-8 input
\usepackage[T1]{fontenc}    % use 8-bit T1 fonts
\usepackage{hyperref}       % hyperlinks
\usepackage{url}            % simple URL typesetting
\usepackage{booktabs}       % professional-quality tables
\usepackage{amsfonts}       % blackboard math symbols
\usepackage{nicefrac}       % compact symbols for 1/2, etc.
\usepackage{microtype}      % microtypography
\usepackage{xcolor}         % colors

\usepackage{graphicx}
\usepackage{multirow}
\usepackage{multicol}
\usepackage{pifont}
\usepackage{wrapfig}
\usepackage{amsmath} 
\usepackage{amssymb}
\usepackage{subfig}
\usepackage{algorithm}
\usepackage{algpseudocode}

\newtheorem{theorem}{Theorem}

\title{Decoupled Kullback-Leibler Divergence Loss}

\author{
	\hspace{-30pt} Jiequan Cui$^{1}$ \hspace{4pt} Zhuotao Tian$^{4}$ \hspace{4pt} Zhisheng Zhong$^{2}$ \hspace{4pt} Xiaojuan Qi$^{3}$ \hspace{4pt} Bei Yu$^{2}$ \hspace{4pt} Hanwang Zhang$^{1}$ \vspace{.3em} \\
	 Nanyang Technological University$^{1}$ \quad\quad\quad
      The Chinese University of Hong Kong$^{2}$ \\
      The University of Hong Kong$^{3}$ \quad\quad\quad HIT(SZ)$^{4}$ \\
	\vspace{-10pt}
}

\begin{document}

\maketitle

\begin{abstract}
In this paper, we delve deeper into the Kullback–Leibler (KL) Divergence loss and mathematically prove that it is equivalent to the Decoupled Kullback-Leibler (DKL) Divergence loss that consists of 1) a weighted Mean Square Error ($\mathbf{w}$MSE) loss and 2) a Cross-Entropy loss incorporating soft labels. 
Thanks to the decomposed formulation of DKL loss, we have identified two areas for improvement. 
Firstly, we address the limitation of KL/DKL in scenarios like knowledge distillation by breaking its asymmetric optimization property. This modification ensures that the $\mathbf{w}$MSE component is always effective during training, providing extra constructive cues.
Secondly, we introduce class-wise global information into KL/DKL to mitigate bias from individual samples.
With these two enhancements, we derive the Improved Kullback–Leibler (IKL) Divergence loss and evaluate its effectiveness by conducting experiments on CIFAR-10/100 and ImageNet datasets, focusing on adversarial training, and knowledge distillation tasks. The proposed approach achieves new state-of-the-art adversarial robustness on the public leaderboard --- \textit{RobustBench} and competitive performance on knowledge distillation, demonstrating the substantial practical merits. Our code is available at \url{https://github.com/jiequancui/DKL}.
\end{abstract}

\section{Introduction}
\label{sec:intro}

Loss functions are a critical component of training deep models. Cross-Entropy loss is particularly important in image classification tasks~\cite{he2016deep, SimonyanZ14a_vgg, DBLP:conf/cvpr/TanCPVSHL19, dosovitskiy2020image, liu2021swin, cui2019fast}, while Mean Square Error (MSE) loss is commonly used in regression tasks~\cite{ren2015faster, he2017mask, he2022masked}. Contrastive loss~\cite{DBLP:conf/icml/ChenK0H20, DBLP:conf/cvpr/He0WXG20, DBLP:journals/corr/abs-2003-04297, DBLP:conf/nips/GrillSATRBDPGAP20, DBLP:conf/nips/CaronMMGBJ20, cui2021parametric, cui2022generalized} has emerged as a popular objective for representation learning. 
The selection of an appropriate loss function can exert a substantial influence on a model's performance. Therefore, the development of effective loss functions~\cite{cao2019learning, focalloss, dkd, wang2020improving, johnson2016perceptual, berman2018lovasz, wen2016discriminative, tan2020equalization, cui2022region} remains a critical research topic in the fields of computer vision and machine learning.

 Kullback-Leibler (KL) Divergence quantifies the degree of dissimilarity between a probability distribution and a reference distribution. As one of the most frequently used loss functions, it finds application in various scenarios, such as adversarial training~\cite{zhang2019theoretically, wu2020adversarial, cui2021learnable, Jia_2022_CVPR}, knowledge distillation~\cite{kd,revkd,dkd}, incremental learning~\cite{chaudhry2018riemannian, lee2017overcoming}, and robustness on out-of-distribution data~\cite{hendrycks2019augmix}. Although many of these studies incorporate KL Divergence loss as part of their algorithms, they may not thoroughly investigate the underlying mechanisms of the loss function. To bridge this gap, our paper aims to elucidate the working mechanism of KL Divergence regarding gradient optimization.

\begin{figure*}[tb!]
    \centering
    \includegraphics[width=.9\linewidth]{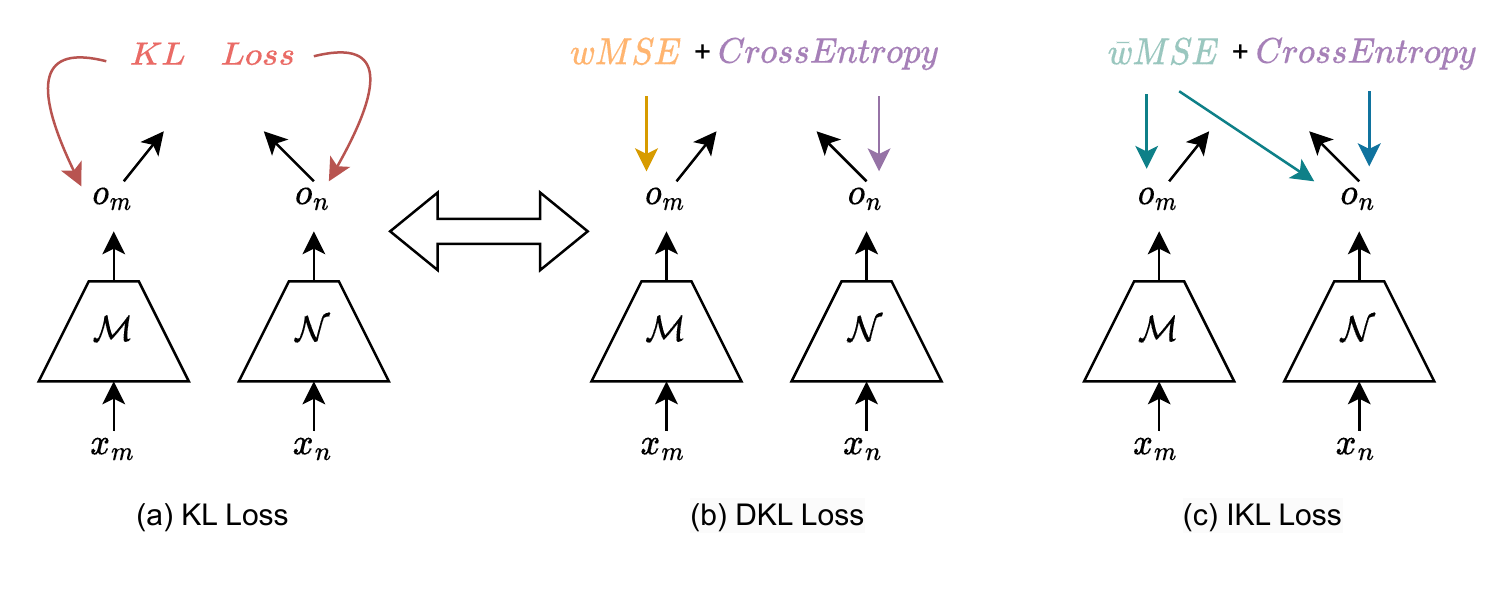}
    \vspace{-0.1in}
    \caption{\textbf{Comparisons of gradient backpropagation between KL, DKL, and IKL losses.} DKL loss is equivalent to KL loss regarding backward optimization.
    $\mathcal{M}$ and $\mathcal{N}$ can be the same one (like in adversarial training) or two separate (like in knowledge distillation) models determined by application scenarios. Similarly, $x_{m}$, $x_{n}$ $\in$ $X$ can also be the same one (like in knowledge distillation) or two different (like in adversarial training) images. $o_{m}$, $o_{n}$ are logits output with which the probability vectors are obtained when applying the \textit{softmax} activation. Black arrows represent the forward process while colored arrows indicate the backward process driven by the corresponding loss functions in the same color. ``$\mathbf{w}$MSE'' is a weighted Mean Square Error (MSE) loss. ``$\mathbf{\bar w}$MSE'' is incorporated with class-wise global information.}
    \label{fig:dkl_ikl}
    \vspace{-0.1in}
\end{figure*}

Our study focuses on the analysis of Kullback–Leibler (KL) Divergence loss from the perspective of gradient optimization. For models with \textit{softmax} activation, we provide theoretical proof that it is equivalent to the Decoupled Kullback–Leibler (DKL) Divergence loss which comprises a weighted Mean Square Error ($\mathbf{w}$MSE) loss and a Cross-Entropy loss with soft labels. Figures~\ref{fig:dkl_ikl}(a) and (b) reveal the equivalence between KL and DKL losses regarding gradient backpropagation. With the decomposed formulation, it becomes more convenient to analyze how the KL loss works in training optimization.
\begin{wrapfigure}{r}{0.38\textwidth}
    \centering
    \includegraphics[width=.90\linewidth]{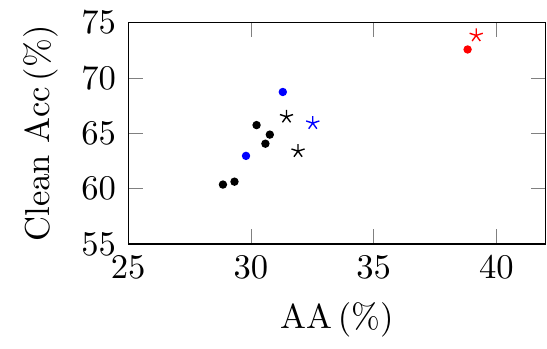}
    \vspace{-0.1in}
    \caption{\textbf{We achieve SOTA robustness on CIFAR-100.} ``star'' represents our method while ``circle'' denotes previous methods. ``Black'' means adversarial training with image preprocessing only including random crop and flip, ``Blue'' is for methods with AutoAug or CutMix, and ``red'' represents methods using synthesized data. AA is short for Auto-Attack~\cite{croce2020reliable}.}
    \label{fig:cifar100_sota}
    \vspace{-0.1in}
\end{wrapfigure}
We have identified potential issues of KL loss with the newly derived DKL loss.
Specifically, its gradient optimization is asymmetric regarding the inputs. As illustrated in Figure~\ref{fig:dkl_ikl}(b), the gradients on $o_{m}$ and $o_{n}$ are asymmetric and driven by the $\mathbf{w}$MSE and Cross-Entropy individually. 
This optimization asymmetry can lead to the $\mathbf{w}$MSE component being ignored in certain scenarios, such as knowledge distillation where $o_{m}$ is the logits of the teacher model and detached from gradient backpropagation.
Fortunately, it is convenient to address this issue with the decoupled formulation of DKL loss by breaking the asymmetric optimization property. As evidenced by Figure~\ref{fig:dkl_ikl}(c), enabling gradient on $o_{n}$ from $\mathbf{w}$MSE alleviates this problem.

Moreover, $\mathbf{w}$MSE component is guided by sample-wise predictions. Hard examples with incorrect prediction scores can lead to challenging optimization. We thus insert class-wise global information to regularize the training process. Integrating DKL with these two points, we derive the Improved Kullback–Leibler (IKL) Divergence loss.

To demonstrate the effectiveness of our proposed IKL loss, we evaluate it with adversarial training and knowledge distillation tasks. Our experimental results on CIFAR-10/100 and ImageNet show that the IKL loss achieves new state-of-the-art robustness on the public leaderboard of \textit{RobustBench} \footnote{https://robustbench.github.io/}. Comparisons with previous methods on adversarial robustness are shown in Figure~\ref{fig:cifar100_sota}.

In summary, the main contributions of our work are:
\begin{itemize} %[leftmargin=15pt]
\item We reveal that the KL loss is mathematically equivalent to a composite of a weighted MSE ($\mathbf{w}$MSE) loss and a Cross-Entropy loss employing soft labels.
\item Based on our analysis, we propose two modifications for enhancement: breaking its asymmetric optimization and incorporating class-wise global information, deriving the Improved Kullback–Leibler (IKL) loss.
\item With the proposed IKL loss, we obtain the state-of-the-art adversarial robustness on \textit{RobustBench} and competitive knowledge distillation performance on CIFAR-10/100 and ImageNet.
\end{itemize}

\section{Related Work}
\label{sec:related_work}
\noindent {\bf Adversarial Robustness.}
Since the identification of adversarial examples by Szegedy et al.~\cite{szegedy2013intriguing}, the security of deep neural networks (DNNs) has gained significant attention,
and ensuring the reliability of DNNs has become a prominent topic in the machine learning community. Adversarial training~\cite{madry2017towards}, being the most effective method, stands out due to its consistently high performance.

Adversarial training incorporates adversarial examples into the training process.
Madary et al.~\cite{madry2017towards} propose the adoption of the universal first-order adversary, specifically the PGD attack, in adversarial training.
Zhang et al.~\cite{zhang2019theoretically} trade off the accuracy and robustness by the KL loss.
Wu et al.~\cite{wu2020adversarial} introduce adversarial weight perturbation to explicitly regulate the flatness of the weight loss landscape.
Cui et al.~\cite{cui2021learnable} leverage guidance from naturally-trained models to regularize the decision boundary in adversarial training.
Additionally, various other techniques~\cite{Jia_2022_CVPR} focusing on optimization or training aspects have also been developed. Besides, recently, several works~\cite{gowal2021improving,wang2023better,addepalli2022efficient} have explored the use of data augmentation techniques to improve adversarial training. We have explored the mechanism of KL loss for adversarial robustness in this paper. The effectiveness of the proposed IKL loss is tested in both settings with and without synthesized data~\cite{karras2022elucidating}.

\noindent {\bf Knowledge Distillation.}
The concept of Knowledge Distillation (KD) was first introduced by Hinton et al.~\cite{kd}.
It involves extracting ``dark knowledge'' from accurate teacher models to guide the learning process of student models. This is achieved by utilizing the KL loss to regularize the output probabilities of student models, aligning them with those of their teacher models when given the same inputs. This simple yet effective technique significantly improves the generalization ability of smaller models and finds extensive applications in various domains. Since the initial success of KD~\cite{kd}, several advanced methods, including logits-based~\cite{cho2019efficacy, furlanello2018born, mirzadeh2020improved, yang2019snapshot, zhang2018deep, dkd, huang2022knowledge} and features-based approaches~\cite{fitnets, crd, ofd, at, revkd, heo2019comprehensive, heo2019knowledge, kim2018paraphrasing, park2019relational, peng2019correlation, yim2017gift}, have been introduced. This paper decouples the KL loss into a new formulation, \textit{i.e.}, DKL, and addresses the limitation of KL loss for application scenarios like knowledge distillation. 

\noindent{\bf Other Applications of KL Divergence Loss.}
In semi-supervised learning, the KL loss acts as a consistency loss between the outputs of weakly and strongly augmented images~\cite{sohn2020fixmatch, tarvainen2017mean}. In continual learning, KL loss helps retain previous knowledge by encouraging consistency between the outputs of pre-trained and newly updated models~\cite{chaudhry2018riemannian, lee2017overcoming}. Additionally, enforcing consistency between outputs from different augmented views enhances model robustness to out-of-distribution data~\cite{hendrycks2019augmix}.

\section{Method}
\label{sec:method}

In this section, we begin by introducing the preliminary mathematical notations in Section~\ref{sec:preliminary}. Theoretical analysis of the equivalence between KL and DKL losses is presented in Section~\ref{sec:dkl}. Finally, we propose the IKL loss to address potential issues of KL/DKL in Section~\ref{sec: IKL}.

\subsection{Preliminary}
\label{sec:preliminary}

\noindent{\bf Definition of KL Divergence}.
Kullback-Leibler (KL) Divergence measures the differences between two probability distributions. For distributions $P$ and $Q$ of a continuous random variable, It is defined to be the integral:
\begin{equation}
   D_{KL}(P||Q) = \int_{-\infty}^{+\infty} p(x) * \log \frac{p(x)}{q(x)} dx, \\
   \label{eq_kl}
\end{equation}
where $p$ and $q$ denote the probability densities of $P$ and $Q$.

The KL loss is one of the most widely used objectives in deep learning, applied across various contexts involving categorical distributions. This paper primarily examines its role in adversarial training and knowledge distillation tasks.

In adversarial training, KL loss enhances model robustness by aligning the output probability distribution of adversarial examples with that of their corresponding clean images, minimizing changes in output despite input perturbations.
In knowledge distillation, KL loss enables a student model to mimic the behavior of a teacher model. Through knowledge transfer from the teacher, the student model is expected to improve its generalization performance.

\noindent{\bf Applications of KL Loss in Deep Learning.}
We consider image classification models that predict probability vectors with the $\textit{softmax}$ activation.
Assume $\mathbf{o}_{i} \in \mathbb{R}^{C}$ is the logits output from one deep model with an image $x_{i} \in X$ as input, where $C$ is the number of classes in the task. $\mathbf{s}_{i} \in \mathbb{R}^{C}$ is the predicted probability vector and $\mathbf{s}_{i} = \textit{softmax}(\mathbf{o}_{i})$. $\mathbf{o}_{i}^{j}$ and $\mathbf{s}_{i}^{j}$ are values for the $j$-th class in $\mathbf{o}_{i}$ and $\mathbf{s}_{i}$ respectively.
KL loss is applied to make $\mathbf{s}_{m}$ and $\mathbf{s}_{n}$ similar in many scenarios, leading to the following objective,
\begin{equation}
   \mathcal{L}_{KL}(x_{m}, x_{n}) =  \sum_{j=1}^{C} \mathbf{s}_{m}^{j} * \log \frac{\mathbf{s}_{m}^{j}}{\mathbf{s}_{n}^{j}}. \\
   \label{eq_kl_sample}
\end{equation}
For instance, in adversarial training, $x_{m}$ is a natural image, and $x_{n}$ is the corresponding adversarial example of $x_{m}$.
Under knowledge distillation settings, $x_{m}$ and $x_{n}$ indicate the same image and are fed into the teacher and student models separately. It is worth noting that $\mathbf{s}_{m}$ is detached from the gradient backpropagation in the knowledge distillation process because the teacher model is well-trained and fixed during training. 

\subsection{Decoupled Kullback-Leibler Divergence Loss}
\label{sec:dkl}
Previous works~\cite{kd, dkd, zhang2019theoretically, cui2021learnable} have incorporated the KL loss into their algorithms without investigating its underlying mechanism. This paper aims to uncover the driving force behind gradient optimization by analyzing the KL loss function.
With the backpropagation rule in training optimization, the derivative gradients are as follows,
\begin{eqnarray}
     \frac{\partial \mathcal{L}_{KL}}{\partial \mathbf{o}_{m}^{j}} &=& \sum_{k=1}^{C} ((\Delta \mathbf{m}_{j,k} -\Delta \mathbf{n}_{j,k}) * (\mathbf{s}_{m}^{k} * \mathbf{s}_{m}^{j})), \label{eq_gradient_1} \\
     \frac{\partial \mathcal{L}_{KL}}{\partial \mathbf{o}_{n}^{j}} &=& \mathbf{s}_{m}^{j} * (\mathbf{s}_{n}^{j} -1 ) + \mathbf{s}_{n}^{j} * (1-\mathbf{s}_{m}^{j}),
     \label{eq_gradient_2}
\end{eqnarray}
where $\Delta \mathbf{m}_{j,k} = \mathbf{o}_{m}^{j} - \mathbf{o}_{m}^{k}$, and $\Delta \mathbf{n}_{j,k} = \mathbf{o}_{n}^{j} - \mathbf{o}_{n}^{k}$.

Leveraging the antiderivative technique alongside the structured gradient information, we introduce a novel formulation called the Decoupled Kullback-Leibler (DKL) Divergence loss, as presented in Theorem~\ref{thm:thm_dkl}. The DKL loss is designed to be equivalent to the KL loss while offering a more analytically tractable alternative for further exploration and study.

\begin{theorem}
	\label{thm:thm_dkl}
	\normalfont{From the perspective of gradient optimization, the Kullback-Leibler (KL) Divergence loss is equivalent to the following Decoupled Kullback-Leibler (DKL) Divergence loss when $\alpha=1$ and $\beta =1$.}
 
        \begin{equation}
            \mathcal{L}_{DKL}(x_m, x_n) = \underbrace{\frac{\alpha}{4} \|\sqrt{\mathcal{S}(\mathbf{w}_{m})}(\Delta \mathbf{m} -\mathcal{S}(\Delta \mathbf{n}))\|^2}_{\textbf{weighted MSE (wMSE)}}  
            \underbrace{-\beta \cdot \mathcal{S}(\mathbf{s}_{m}^{\top}) \cdot \log \mathbf{s}_{n}}_{\textbf{Cross-Entropy}},
        \label{eq_dkl}
        \end{equation}
        where $\mathcal{S}(\cdot)$ represents \textit{stop gradients} operation,
        $\mathbf{s}_{m}^{\top}$ is transpose of $\mathbf{s}_{m}$,
        $\mathbf{w}_{m}^{j,k}$ = $\mathbf{s}_{m}^{j} * \mathbf{s}_{m}^{k}$,
        $\Delta \mathbf{m}_{j,k} = \mathbf{o}_{m}^{j} - \mathbf{o}_{m}^{k}$, 
        and $\Delta \mathbf{n}_{j,k} = \mathbf{o}_{n}^{j} - \mathbf{o}_{n}^{k}$.
        Summation is used for the reduction of $\|\cdot\|^2$.

        \textit{Proof~~} See Appendix~\ref{sec:proof}.
\end{theorem}

\noindent {\bf Interpretation}.
With Theorem~\ref{thm:thm_dkl}, we know that KL loss is equivalent to DKL loss regarding gradient optimization, \textit{i.e., DKL loss produces the same gradients as KL loss given the same inputs.} Therefore, KL loss can be interpreted as a composition of a $\mathbf{w}$MSE loss and a Cross-Entropy loss. This is the first work to reveal the precise quantitative relationships between KL, Cross-Entropy, and MSE losses. Upon examining this new formulation, we identify two potential issues with the KL loss.

{\bf Asymmetirc Optimization.} As shown in Eqs.~\eqref{eq_gradient_1} and~\eqref{eq_gradient_2}, gradient optimization is asymmetric for $\mathbf{o}_{m}$ and $\mathbf{o}_{n}$.
The $\mathbf{w}$MSE and Cross-Entropy losses in Theorem~\ref{thm:thm_dkl} are complementary and collaboratively work together to make $\mathbf{o}_{m}$ and $\mathbf{o}_{n}$ similar. 
Nevertheless, the asymmetric optimization can cause the $\mathbf{w}$MSE component to be neglected or overlooked when $\mathbf{o}_{m}$ is detached from gradient backpropagation, which is the case for knowledge distillation, potentially leading to performance degradation.

{\bf Sample-wise Prediction Bias.} As shown in Eq.~\eqref{eq_dkl}, $\mathbf{w}_{m}$ in $\mathbf{w}$MSE component is conditioned on the prediction score of $x_{m}$. However, sample-wise predictions can be subject to significant variance. Incorrect prediction of hard examples or outliers will mislead the optimization and result in unstable training. Our study in Sections~\ref{sec:case_study} and ~\ref{sec:ablation} indicates that the choice of $\mathbf{w}_{m}$ significantly affects adversarial robustness. 

\vspace{-0.1in}
\subsection{Improved Kullback-Leibler Divergence Loss}
\label{sec: IKL}
Based on the analysis in Section~\ref{sec:dkl}, we propose an Improved Kullback-Leibler (IKL) Divergence loss.
Distinguished from DKL in Theorem~\ref{thm:thm_dkl}, we make the following improvements: 1) \textit{breaking the asymmetric optimization property}; 2) \textit{inserting class-wise global information to mitigate sample-wise bias}. The details are presented as follows.

\noindent{\bf Breaking the Asymmetric Optimization Property}.
As shown in Eq.~\eqref{eq_dkl}, the $\mathbf{w}$MSE component encourages $\mathbf{o}_{m}$ to resemble $\mathbf{o}_{n}$ by capturing second-order information, specifically the differences between logits for each pair of classes. Each addend in $\mathbf{w}$MSE only involves logits of two classes. We refer to this property as \textit{locality}.
On the other hand, the Cross-Entropy loss ensures that $\mathbf{s}_{n}$ and $\mathbf{s}_{m}$ produce similar predicted scores. Each addend in the Cross-Entropy gathers all class logits. We refer to this property as \textit{globality}.
Two loss terms collaboratively work together to make $\mathbf{o}_{n}$ and $\mathbf{o}_{m}$ similar in \textit{locality} and \textit{globality}. Discarding any one of them can lead to performance degradation.

However, because of the asymmetric optimization property of KL/DKL, the unexpected case can occur when $\mathbf{s}_{m}$ is detached from the gradient backpropagation (scenarios like knowledge distillation), in which the formulation will be:

\begin{equation}
    \mathcal{L}_{DKL-KD}(x_m, x_n) = \underbrace{\frac{\alpha}{4} \|\sqrt{\mathcal{S}(\mathbf{w}_{m})}(\mathcal{S}(\Delta \mathbf{m}) - \mathcal{S}(\Delta \mathbf{n}))\|^2}_{\textbf{weighted MSE (wMSE)}} 
    \underbrace{-\beta \cdot \mathcal{S}(\mathbf{s}_{m}^{\top}) \cdot \log \mathbf{s}_{n}}_{\textbf{Cross-Entropy}}
    \label{eq:eq_dkl_distill_1}
\end{equation}

As indicated by Eq.~\eqref{eq:eq_dkl_distill_1}, the $\mathbf{w}$MSE component loss takes no effect on training optimization since all sub-components of $\mathbf{w}$MSE are detached from gradient propagation, which can potentially hurt the model performance. Knowledge distillation exactly matches this case because the teacher model is fixed during knowledge distillation training.
Thanks to the decomposition of DKL formulation, we address this issue by breaking the asymmetric optimization property, \textit{i.e.}, enabling the gradients of $\mathcal{S}(\Delta \mathbf{n})$ in Eq.~\eqref{eq_dkl}. Then, the updated formulation of Eq.~\eqref{eq:eq_dkl_distill_1} becomes,

\begin{eqnarray}
    \widehat{\mathcal{L}}_{DKL-KD}(x_m, x_n) = \underbrace{\frac{\alpha}{4} \|\sqrt{\mathcal{S}(\mathbf{w}_{m})}(\mathcal{S}(\Delta \mathbf{m}) - \Delta \mathbf{n})\|^2}_{\textbf{weighted MSE (wMSE)}}  
    \underbrace{-\beta \cdot \mathcal{S}(\mathbf{s}_{m}^{\top}) \cdot \log \mathbf{s}_{n}}_{\textbf{Cross-Entropy}}.
    \label{eq_dkl_distill_2}
\end{eqnarray}

After enabling the gradients of $\mathcal{S}(\Delta \mathbf{n})$ in Eq.~\eqref{eq_dkl}, $\mathbf{w}$MSE will produce symmetric gradients on $o_{n}$ and $o_{m}$. Regarding the knowledge distillation, $\mathbf{w}$MSE can output gradient on $o_{n}$ and promote the training optimization demonstrated by Eq.~\eqref{eq_dkl_distill_2}.

\noindent{\bf Inserting Class-wise Global Information}.
Recall in Theorem~\ref{thm:thm_dkl}, $\mathbf{w}_{m}$ in Eq.~\eqref{eq_dkl} is calculated as: 
\begin{equation}
        \mathbf{w}_{m}^{j,k} = \mathbf{s}_{m}^{j} * \mathbf{s}_{m}^{k}.
\end{equation}
It indicates that $\mathbf{w}_{m}$ depends on the sample-wise prediction scores. Nevertheless, the model cannot output correct predictions when dealing with outliers or hard examples in training. In this case, $\mathbf{w}$MSE will attach the most importance on the predicted class $\hat{y} = \arg \max \mathcal{M}(x_{m})$ rather than the ground-truth class, which misleads the optimization and makes the training unstable.

We thus insert class-wise global information into $\mathbf{w}$MSE component, replacing $\mathbf{w}_{m}$ with $\mathbf{\bar w}_{y}$:
\begin{equation}
        \mathbf{\bar w}_{y}^{j,k} = \mathbf{\bar s}_{y}^{j} * \mathbf{\bar s}_{y}^{k},
        \label{eq:class_weight}
\end{equation}
where $y$ is ground-truth label of $x_{m}$, $\mathbf{\bar s}_{y}=\frac{1}{|X_{y}|} \sum_{x_{i} \in X_{y}} \mathbf{s}_{i}$.

The class-wise global information injected by $\mathbf{\bar w}_{y}$ can act as a regularization to enhance intra-class consistency and mitigate biases that may arise from sample noises. Especially, in the late stage of training, $\mathbf{\bar w}_{y}$ can always provide correct predictions, benefiting the optimization of $\mathbf{\bar w}$MSE component.  

To this end, we derive the IKL loss in Eq.~\eqref{eq_ikl} by incorporating the two designs,
\begin{eqnarray}
    \mathcal{L}_{IKL}(x_m, x_n) =& \underbrace{\frac{\alpha}{4} \|\sqrt{\mathcal{S}(\mathbf{\bar w}_{y})}(\Delta \mathbf{m}-\Delta \mathbf{n})\|^2}_{\textbf{weighted MSE ($\mathbf{\bar w}$MSE)}}  
    \underbrace{-\beta \cdot \mathcal{S}(\mathbf{s}_{m}^{\top}) \cdot \log \mathbf{s}_{n}}_{\textbf{Cross-Entropy}},
    \label{eq_ikl}
\end{eqnarray}
where $y$ is the ground-truth label for $x_{m}$, $\mathbf{\bar w}_{y} \in \mathbb{R}^{C \times C}$ is the weights for class $y$ calculated with Eq.~\eqref{eq:class_weight}.

\begin{table*}[tb!]
    \centering
    %\small
    \caption{\textbf{Ablation study on ``GI'' and ``BA'' with DKL loss.} ``GI'' represents ``Inserting Global Information'', and ``BA'' indicates ``Breaking Asymmetric Optimization''. ``Clean'' is the test accuracy of clean images and ``AA'' is the robustness under Auto-Attack. CIFAR-100 is used for the adversarial training task and ImageNet is adopted for the knowledge distillation task.}
    \vspace{-0.1in}
    \resizebox{1.0\linewidth}{!}
    {
    \begin{tabular}{lcccccc}
    \toprule
         \textbf{Index}    &\textbf{GI} & \textbf{BA} &\multicolumn{2}{c}{\textbf{Adversarial Training}} &\textbf{Knowledge Distillation} &\textbf{Descriptions} \\
         & & & \textbf{Clean (\%)} & \textbf{AA (\%)} & \textbf{Top-1 (\%)} & \\
        \midrule
        (a)           &Na        & Na       &62.87 &30.29  &71.03 &baseline with KL loss.   \\  
        \midrule
        (b)           &\ding{55} &\ding{55} &62.54 &30.20           &71.03          &DKL, equivalent to KL loss.  \\
        (c)           &\ding{55} &\ding{52} &62.69 &30.42           &71.80 &(b) with BA.        \\
        (d)           &\ding{52} &\ding{52} &65.76 &\textbf{31.91}  &\textbf{71.91} &(c) with GI, \textit{i.e.}, IKL. \\
        \bottomrule
    \end{tabular} 
    }
    \label{tab:global_info_dkl}
    \vspace{-0.1in}
\end{table*}

\begin{figure*}[tb!]
    \centering
    \subfloat[IKL-AT]            { \includegraphics[width=0.308\linewidth]{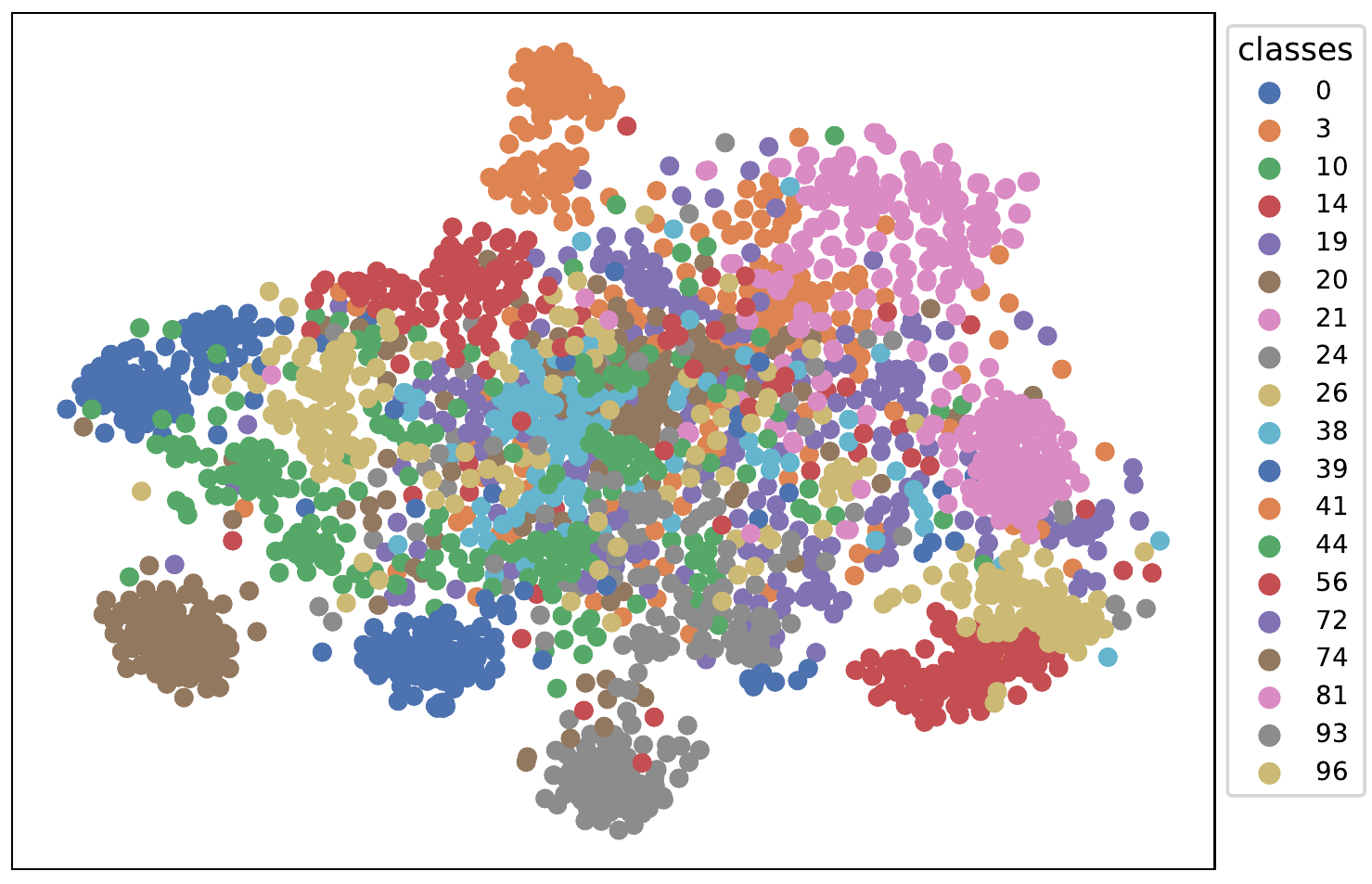} \label{fig:tsne_ikl_at}}
    \subfloat[TRADES]            { \includegraphics[width=0.308\linewidth]{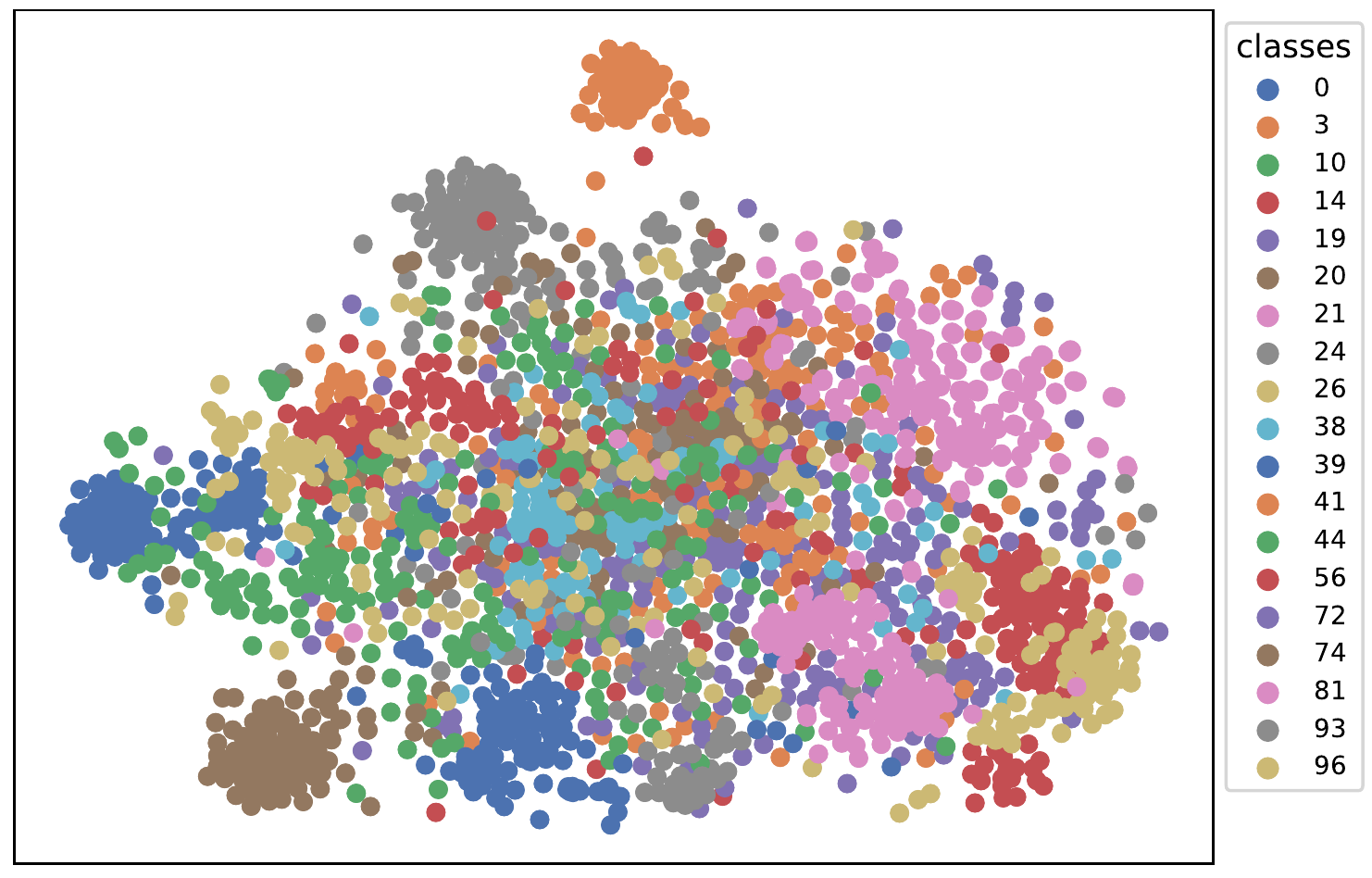}  \label{fig:tsne_trades}}
    \subfloat[Margin differences]{ \includegraphics[width=0.308\linewidth]{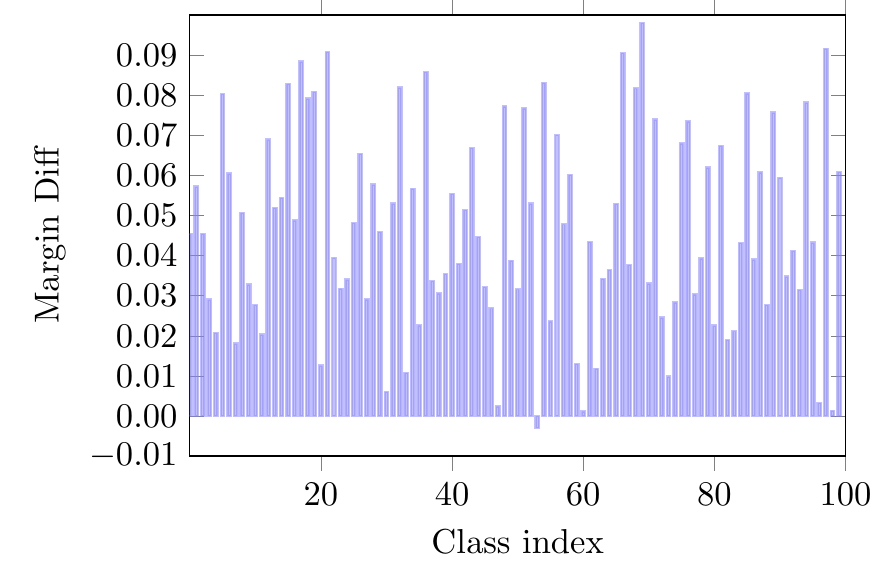}        \label{fig:margin_diff}}
    \caption{
        \textbf{Visualization comparisons.}
        (a) t-SNE visualization of the model trained by IKL-AT on CIFAR-100;
        (b) t-SNE visualization of the model trained by TRADES on CIFAR-100.
        (c) Class margin differences between models trained by IKL-AT and TRADES.
    }
\end{figure*}

\subsection{A Case Study and Analysis}
\label{sec:case_study}
\noindent {\bf A Case Study}.
We empirically examine each component of IKL on CIFAR-100 with the adversarial training task and on ImageNet with the knowledge distillation task.
Ablation experimental results and their setting descriptions are listed in Table~\ref{tab:global_info_dkl}.
In the implementation, for adversarial training, we use improved TRADES~\cite{zhang2019theoretically} as our baseline that combines with AWP~\cite{wu2020adversarial} and uses an increasing epsilon schedule~\cite{addepalli2022efficient}. For knowledge distillation, we use the official code from DKD.
The comparison between (a) and (b) shows that DKL can achieve comparable performance, confirming the equivalence to KL. 
The comparisons among (b), (c), and (d) confirm the effectiveness of the ``GI'' and ``BA".

\noindent {\bf Analysis on Inserting Class-wise Global Information}.
As evidenced by Table~\ref{tab:global_info_dkl}, class-wise global information plays an important role in adversarial robustness.
The mean probability vector $\bar s_{y}$ of all samples in the class $y$ is more robust than the sample-wise probability vector.
During training, once the model gives incorrect predictions for hard samples or outliers,
$\mathbf{w}_{m}$ in Eq.~\eqref{eq_dkl} will wrongly guide the optimization.
Adoption of $\mathbf{\bar w}_{y}$ in Eq.~\eqref{eq_ikl} can mitigate the issue and meanwhile enhance intra-class consistency.

To visualize the effectiveness of inserting class-wise global information, we define the boundary margin for class $y$ as:
\begin{equation}
    \text{Margin}_{y} = \bar s_{y} [y] - \max_{k \neq y} \bar s_{y} [k].
\end{equation}

We plot the margin differences between models trained by IKL-AT and TRADES on CIFAR-100.
As shown in Figure~\ref{fig:margin_diff}, almost all class margin differences are positive, demonstrating that there are larger decision boundary margins for the IKL-AT model.
Such larger margins lead to stronger robustness. This phenomenon is coherent with our experimental results in Section~\ref{sec:adv}.

We also randomly sample 20 classes in CIFAR-100 for t-SNE visualization. The numbers in the pictures are class indexes. For each sampled
class, we collect the feature representation of natural images and adversarial examples with the validation set.
The visualization by t-SNE is shown in Figures~\ref{fig:tsne_trades} and~\ref{fig:tsne_ikl_at}.
Compared with TRADES that trained with KL loss, features by IKL-AT models are more compact and separable.

\section{Experiments}
\label{sec:exp}

To verify the effectiveness of our IKL loss, we conduct experiments on CIFAR-10, CIFAR100, and ImageNet for adversarial training (Section~\ref{sec:adv}) and knowledge distillation (Sections~\ref{sec:kd} and~\ref{sec:imbalanced_kd}).

\subsection{Adversarial Robustness}
\label{sec:adv}
\noindent {\bf Experimental Settings}.
We use an improved version of TRADES~\cite{zhang2019theoretically} as our baseline, which incorporates AWP~\cite{wu2020adversarial} and adopts an increasing epsilon schedule. SGD optimizer with a momentum of 0.9 is used. We use the cosine learning rate strategy with an initial learning rate of 0.2 and train models 200 epochs. The batch size is 128, the weight decay is 5e-4 and the perturbation size $\epsilon$ is set to 8/255. Following previous work~\cite{zhang2019theoretically, cui2021learnable}, standard data augmentation including random crops and random horizontal flip is performed for data preprocessing. Models are trained with 4 Nvidia GeForce 3090 GPUs.

Under the setting of training with synthesized data by generative models, we strictly follow the training configurations in DM-AT~\cite{wang2023better} for fair comparisons. Our implementations are based on their open-sourced code. We only replace the KL loss with our IKL loss.

\begin{table*}[tb!]
    \centering
    \caption{
        \textbf{Test accuracy (\%) of clean images and robustness (\%) under AutoAttack on CIFAR-100.}
        All results are the average over three trials.
    }
    \vspace{-0.1in}
    \label{tab:sota_cifar100}
    %{{{
    %\resizebox{.918\linewidth}{!}
    {
        \begin{tabular}{cclccc}
            \toprule
            Dataset  & Method & Architecture & Augmentation Type & Clean & AA \\
            \midrule
            \multirow{10}{*}{\begin{tabular}{c}\textbf{CIFAR-100}\\
                ($\ell_{\infty}$, $\epsilon=8/255$)
            \end{tabular}}  & AWP      & WRN-34-10     & Basic        & 60.38 & 28.86 \\
            & LBGAT    & WRN-34-10     & Basic        & 60.64 & 29.33 \\
            & LAS-AT   & WRN-34-10     & Basic        & 64.89  & 30.77 \\
            & ACAT     & WRN-34-10     & Basic        & 65.75  & 30.23 \\
            & \textbf{IKL-AT}          & WRN-34-10    & Basic        & \textbf{65.76} & \textbf{31.91} \\
            \cmidrule{2-6}
            & ACAT      & WRN-34-10     & AutoAug        & \textbf{68.74}  & 31.30 \\
            & \textbf{IKL-AT}                         & WRN-34-10     & AutoAug        &66.08  & \textbf{32.53} \\
            \cmidrule{2-6}
            &DM-AT~\cite{wang2023better} & WRN-28-10     & 50M Generated Data    & 72.58 & 38.83 \\
            & \textbf{IKL-AT}       & WRN-28-10     & 50M Generated Data    & \textbf{73.85} &\textbf{39.18} \\
            \bottomrule
        \end{tabular}%
    }
    %}}}
%\vspace{-0.05in}    
\end{table*}%

\begin{table*}[t]
\centering
%\small
%\footnotesize
%\setlength{\tabcolsep}{2.2pt}
\caption{
    \textbf{Test accuracy (\%) of clean images and robustness (\%) under AutoAttack on CIFAR-10.} Average over three trials are listed.
}
\vspace{-0.1in}
%\resizebox{.916\linewidth}{!}
{
\renewcommand*{\arraystretch}{1.0}
\begin{tabular}{cclccc}
\toprule
Dataset  & Method & Architecture & Augmentation Type & Clean & AA \\
\midrule
\multirow{10}{*}{\begin{tabular}{c}\textbf{CIFAR-10} \\
    ($\ell_{\infty}$, $\epsilon=8/255$)
                \end{tabular}}        & Rice et al.~\cite{rice2020overfitting}           & WRN-34-20       & Basic  & 85.34 & 53.42 \\
                                      & LBGAT        & WRN-34-20       & Basic  & \textbf{88.70} & 53.57 \\
                                      & AWP          & WRN-34-10       & Basic  & 85.36 & 56.17 \\ 
                                      & LAS-AT       & WRN-34-10       & Basic  & 87.74 & 55.52 \\
                                      & ACAT         & WRN-34-10       & Basic  & 82.41 & 55.36 \\
                                      & \textbf{IKL-AT}                      & WRN-34-10       & Basic  & 84.80 & \textbf{57.09} \\
                                      \cmidrule{2-6}
                                      & ACAT    & WRN-34-10       & AutoAug  & 88.64 & 57.05 \\
                                      & \textbf{IKL-AT}                      & WRN-34-10       & AutoAug  & 85.20 & \textbf{57.62} \\
                                      \cmidrule{2-6}
                                   
                                      & DM-AT~\cite{wang2023better}                & WRN-28-10       & 20M Generated Data    & 92.44 & 67.31 \\ 
                                      %\cmidrule{2-6}
                                      & \multirow{1}{*}{\textbf{IKL-AT}}     & WRN-28-10       & 20M Generated Data   &92.16 &\textbf{67.75} \\
\bottomrule
\end{tabular}%
\label{tab:sota_cifar10}
}
\vspace{-0.1in}
\end{table*}%

\noindent {\bf Datasets and Evaluation}.
Following previous work~\cite{wu2020adversarial, cui2021learnable}, CIFAR-10 and CIFAR-100 are used for the adversarial training task. we report the clean accuracy on natural images and adversarial robustness under Auto-Attack~\cite{croce2020reliable} with epsilon 8/255. 

\begin{table*}[tb!]
    \center
    %\small
    \caption{\textbf{Top-1 accuracy~(\%) on the ImageNet validation and training speed (sec/iteration) comparisons.} Training speed is calculated on 4 Nvidia GeForce 3090 GPUs with a batch of 512 224x224 images.
    All results are the average over three trials.
    }
    \vspace{-0.1in}
%\resizebox{.95\linewidth}{!}
{
\begin{tabular}{ccccccc}
\toprule
\multirow{4}{*}{\begin{tabular}[c]{@{}c@{}}Distillation \\ Manner\end{tabular}} & \multirow{2}{*}{Teacher}  & \multirow{4}{*}{Extra Parameters} & \multicolumn{2}{c}{ResNet34}  & \multicolumn{2}{c}{ResNet50} \\
&      & & \multicolumn{2}{c}{73.31} &\multicolumn{2}{c}{76.16}   \\
&  \multirow{2}{*}{Student}  & & \multicolumn{2}{c}{ResNet18} & \multicolumn{2}{c}{MobileNet} \\
& \space     & & \multicolumn{2}{c}{69.75} & \multicolumn{2}{c}{68.87} \\
\midrule
\multirow{4}{*}{Features}
& AT             & \ding{55} & 70.69   &                 & 69.56    &\\
& OFD            & \ding{52} & 70.81   &                 & 71.25    &\\
& CRD            & \ding{52} & 71.17   &                 & 71.37    &\\ 
& ReviewKD       & \ding{52} & 71.61   &0.319 s/iter     & 72.56    &0.526 s/iter\\ 
\midrule
\multirow{4}{*}{Logits}   
& DKD         & \ding{55} & 71.70                            &                 & 72.05 &\\
& KD          & \ding{55} & 71.03                            &                 & 70.50 &\\
& \textbf{IKL-KD}      & \ding{55} & \textbf{71.91}                   &\textbf{0.197 s/iter}     & \textbf{72.84}                 &\textbf{0.252 s/iter}  \\
%& $\Delta$             &        & \textcolor{myorange}{+0.88}      &              &\textcolor{myorange}{+2.34}     & \\
\bottomrule
\end{tabular}
}
\label{tab:imagenet_kd}
\vspace{-0.1in}
\end{table*}

\noindent {\bf Comparison Methods}.
To compare with previous methods, we categorize them into two groups according to the different types of data preprocessing:
\begin{itemize}%[leftmargin=15pt]
    \item Methods with basic augmentation, \textit{i.e.}, random crops and random horizontal flip.
    \item Methods using augmentation with generative models or Auto-Aug~\cite{DBLP:conf/cvpr/CubukZMVL19}, CutMix~\cite{yun2019cutmix}. 
\end{itemize}

\noindent {\bf Comparisons with State-of-the-art on CIFAR-100}.
On CIFAR-100, with the basic augmentations setting, we compare with AWP, LBGAT, LAS-AT, and ACAT.
The experimental results are summarized in Table~\ref{tab:sota_cifar100}. Our WRN-34-10 models trained with IKL loss do a better trade-off between natural accuracy and adversarial robustness. With $\frac{\alpha}{4}=5$ and $\beta=5$, the model achieves \textbf{65.76\%} top-1 accuracy on natural images while \textbf{31.91\%} adversarial robustness under Auto-Attack. 
An interesting phenomenon is that IKL-AT is complementary to data augmentation strategies, like AutoAug, without any specific designs, which is different from the previous observation that the data augmentation strategy hardly benefits adversarial training~\cite{wu2020adversarial}. 
With AutoAug, we obtain \textbf{32.53\%} adversarial robustness, achieving new state-of-the-art under the setting without extra real or generated data.  

We follow DM-AT~\cite{wang2023better} to take advantage of synthesized images generated by the popular diffusion models~\cite{karras2022elucidating}.  
With 50M generated images, we create new state-of-the-art with WideResNet-28-10, achieving \textbf{73.85\%} top-1 natural accuracy and \textbf{39.18\%} adversarial robustness under Auto-Attack.

\noindent {\bf Comparison with State-of-the-art on CIFAR-10}.
Experimental results on CIFAR-10 are listed in Table~\ref{tab:sota_cifar10}. With the basic augmentation setting,
our model achieves 84.80\% top-1 accuracy on natural images and 57.09\% robustness,
outperforming AWP by 0.92\% on robustness.
With extra generated data, we improve the state-of-the-art by 0.44\%, achieving \textbf{67.75\%} robustness.

\subsection{Knowledge Distillation}
\label{sec:kd}
\noindent {\bf Datasets and Evaluation}.
Following previous work~\cite{revkd,crd}, we conduct experiments on CIFAR-100~\cite{cifar} and ImageNet~\cite{imagenet} to show the advantages of IKL on knowledge distillation. %ImageNet~\cite{imagenet} is the most challenging dataset for classification, which consists of 1.2 million images for training and 50K images for validation over 1,000 classes.
For evaluation, we report top-1 accuracy on CIFAR-100 and ImageNet validation. The training speed of different methods is also discussed.

\noindent {\bf Experimental Settings}.
We follow the experimental settings in DKD. Our implementation for knowledge distillation is based on their open-sourced code. Models are trained with 1 and 8 Nvidia GeForce 3090 GPUs on CIFAR and ImageNet separately. 

Specifically, on CIFAR-100, we train all models for 240 epochs with a learning rate that decayed by 0.1 at the 150th, 180th, and 210th epoch. We initialize the learning rate to 0.01 for MobileNet and ShuffleNet, and 0.05 for other models. The batch size is 64 for all models. We train all models three times and report the mean accuracy. 
On ImageNet, we use the standard training that trains the model for 100 epochs and decays the learning rate for every 30 epochs. We initialize the learning rate to 0.2 and
set the batch size to 512.

For both CIFAR-100 and ImageNet, we consider the distillation among the architectures having the same unit structures, like ResNet56 and ResNet20, VGGNet13 and VGGNet8. On the other hand, we also explore the distillation among architectures made up of different unit structures, like WideResNet and ShuffleNet, VggNet and MobileNet-V2.

\noindent {\bf Comparison Methods}.
According to the information extracted from the teacher model in distillation training, knowledge distillation methods can be divided into two categories: 
\begin{itemize} %[leftmargin=15pt]
    \item Features-based methods~\cite{fitnets,crd, revkd, ofd}. This kind of method makes use of features from different layers of the teacher model, which can need extra parameters and high training computational costs.
    \item Logits-based methods~\cite{kd, dkd}. This kind of method only makes use of the logits output of the teacher model, which does not require knowing the architectures of the teacher model and thus is more general in practice.
\end{itemize}

\noindent {\bf Comparison with State-of-the-art on CIFAR-100}.
Experimental results on CIFAR-100 are summarized in Table~\ref{tab:cifar_kd} and Table~\ref{tab:cifar2_kd} (in Appendix).
Table~\ref{tab:cifar_kd} lists the comparisons with previous methods under the setting that the architectures of the teacher and student have the same unit structures. Models trained by IKL-KD can achieve comparable or better performance in all considered settings. Specifically, we achieve the best performance in 4 out of 6 training settings.
Table~\ref{tab:cifar2_kd} shows the comparisons with previous methods under the setting that the architectures of the teacher and student have different unit structures. We achieve the best performance in 3 out of 5 training configurations.

\noindent {\bf Comparison with State-of-the-art on ImageNet}.
We empirically show the comparisons with other methods on ImageNet in Table~\ref{tab:imagenet_kd}.
With a ResNet34 teacher, our ResNet18 achieves \textbf{71.91\%} top-1 accuracy. With a ResNet50 teacher, our MobileNet achieves \textbf{72.84\%} top-1 accuracy. Models trained by IKL-KD surpass all previous methods while saving \textbf{38\%} and \textbf{52\%} computation costs for ResNet34--ResNet18 and ResNet50--MobileNet distillation training respectively when compared with ReviewKD~\cite{revkd}.

\subsection{Knowledge Distillation on Imbalanced Data}
\label{sec:imbalanced_kd}
Data often follows a long-tailed distribution. Tackling the long-tailed recognition problem is essential for real-world applications. Lots of research has contributed to algorithms and theories~\cite{cao2019learning, cui2019class, kang2019decoupling, menon2020long,cui2022generalized,cui2021parametric, cui2024classes} on the problem.
In this work, we examine how the knowledge distillation with our IKL loss affects model performance on imbalanced data, \textit{i.e.}, ImageNet-LT~\cite{liu2019large}. We train ResNets models 90 epochs with \textit{Random-Resized-Crop} and horizontal flip as image pre-processing. Following previous work~\cite{9774921}, we report the top-1 accuracy on Many-shot, Meidum-shot, and All classes. As shown in Table~\ref{tab:imagenetlt_kd}, IKL-KD consistently outperforms KL-KD on imbalanced data.

\begin{table*}[tb!]
    \center
    \caption{\textbf{Peformance~(\%) on imbalanced data, \textit{i.e.}, the ImageNet-LT.}}
    \vspace{-0.1in}
%\resizebox{.95\linewidth}{!}
{
\begin{tabular}{ccccccc}
\toprule
Method   & Teacher &Student  &Many(\%) &Medium(\%) &Few(\%) &All(\%)\\
\midrule
Baseline & - &ResNet-18  &63.16 &33.47 &5.88 &41.15 \\
Baseline & - &ResNet-50  &67.25 &38.56 &8.21 &45.47 \\
Baseline & - &ResNet-101 &68.91 &42.32 &11.24 &48.33 \\
\midrule
KL-KD    &ResNeXt-101  &ResNet-18 &64.6 &37.88 &9.53  &44.32 \\
KL-KD    &ResNeXt-101  &ResNet-50 &68.83 &42.31 &11.37 &48.31 \\
\midrule
\textbf{IKL-KD}    &ResNeXt-101  &ResNet-18 &66.60 &38.53 &8.19  &\textbf{45.21} \\
\textbf{IKL-KD}    &ResNeXt-101  &ResNet-50 &70.06 &43.47 &10.99 &\textbf{49.29} \\
\bottomrule
\end{tabular}
}
\label{tab:imagenetlt_kd}
\vspace{-0.1in}
\end{table*}

\subsection{Ablation Studies}
\label{sec:ablation}
\noindent {\bf Ablation on $\alpha$ and $\beta$ for Adversarial Robustness}.
Thanks to the decomposition of the DKL loss formulation, the two components ($\mathbf{w}$MSE and Cross-Entropy) of IKL can be manipulated independently. We empirically study the effects of hyper-parameters of $\alpha$ and $\beta$ on CIFAR-100 for adversarial robustness. Clean accuracy on natural data and robustness under AA~\cite{croce2020reliable} are reported in Table~\ref{tab:ablation_alpha} and Table~\ref{tab:ablation_beta}. Reasonable $\alpha$ and $\beta$ should be chosen for the best trade-off between natural accuracy and adversarial robustness.

\begin{table*}[tb!]
\centering
\caption{\textbf{Ablation study on hyper-parameters of IKL.}}
\vspace{-0.1in}
\begin{minipage}{0.32\textwidth}
\centering
%\resizebox{1.0\linewidth}{!}
{
\begin{tabular}{ccc}
\toprule
$\frac{\alpha}{4}$ & Clean  & AA\\
\midrule
    $\frac{\alpha}{4}=3$     & 67.52 &31.29 \\
    $\frac{\alpha}{4}=4$     & 66.26 &31.33 \\ 
    $\frac{\alpha}{4}=5$     & 65.76 &31.91 \\ 
    $\frac{\alpha}{4}=6$     & 65.14 &31.64 \\ 
\bottomrule
\end{tabular}
}
\caption{Effects of $\frac{\alpha}{4}$.}
\label{tab:ablation_alpha}
\end{minipage}
%\hspace{0.01in}
\begin{minipage}{0.32\textwidth}  
\centering
%\resizebox{1.0\linewidth}{!}
{
\begin{tabular}{ccc}
\toprule
\textbf{$\beta$} & Clean & AA \\
\midrule
    $\beta=2$     & 66.13 &30.95 \\
    $\beta=3$     & 66.31 &31.33 \\
    $\beta=4$     & 66.00 &31.57 \\  
    $\beta=5$     & 65.76 &31.91 \\ 
\bottomrule
\end{tabular}
}
\caption{Effects of $\beta$.}
\label{tab:ablation_beta}
\end{minipage}
%\hspace{0.01in}
\begin{minipage}{0.32\textwidth}  
\centering
%\resizebox{1.0\linewidth}{!}
{
\begin{tabular}{ccc}
\toprule
\textbf{$\tau$} & Clean & AA \\
\midrule
    $\tau=1$     & 59.99 &31.35 \\
    $\tau=2$     & 63.77 &31.88 \\
    $\tau=3$     & 65.28 &31.69 \\  
    $\tau=4$     & 65.76 &31.91 \\ 
\bottomrule
\end{tabular}
}
\caption{Effects of $\tau$.}
\label{tab:ablation_tau}
\end{minipage}
\label{tab:ablation_alpha_beta_tau}
\vspace{-0.25in}
\end{table*}

\noindent {\bf Ablation on Temperature ($\tau$) for Global Information.}
As discussed in Section~\ref{sec: IKL},  the incorporated class-wise global information is proposed to promote intra-class consistency and mitigate the biases from sample noises.
When calculating the $\bar w_{y}$ and $\bar s_{y}$, a temperature $\tau$ could be applied before getting sample probability vectors.
We summarize the experimental results in Table~\ref{tab:ablation_tau} for ablation of $\tau$. Interestingly, we observe that models usually exhibit higher performance on clean images with a higher $\tau$. There are even 5.75\% improvements of clear accuracy while keeping comparable robustness when changing $\tau=1$ to $\tau=4$, which implies the vast importance of weights in $\mathbf{w}$MSE component of DKL/KL for adversarial robustness.
To achieve the strongest robustness, we finally choose $\tau=4$ as illustrated by empirical study. 

\noindent {\bf Ablation on Various Perturbation Size $\epsilon$}.
We evaluate model robustness with unknown perturbation size $\epsilon$ in training under Auto-Attack.
The experimental results are summarized in Table~\ref{tab:ablation_epsilon}.
As shown in Table~\ref{tab:ablation_epsilon}, model robustness decreases significantly as the $\epsilon$ increases for both the TRADES model and our model. Nevertheless, our model achieves stronger robustness than the TRADES model under all of $\epsilon$, outperforming TRADES by 1.34\% on average robustness.
The experimental results demonstrate the super advantages of models adversarially trained with our IKL loss.

\begin{table}[tb!]
\centering
\caption{\textbf{Ablation study of $\epsilon$.}}
%\resizebox{0.7\linewidth}{!}
{
\setlength{\tabcolsep}{2.9mm}
\begin{tabular}{ccccccccc}
\toprule
\multirow{2}{*}{Method} &\multirow{2}{*}{Clean}  &\multicolumn{7}{c}{AA} \\
\cmidrule{3-9} 
& &$\frac{2}{255}$ &$\frac{4}{255}$ & $\frac{6}{255}$ & $\frac{8}{255}$ & $\frac{10}{255}$ &$\frac{12}{255}$ &Avg.\\
\midrule
   TRADES &62.87 &53.88 &45.31 &37.28 &30.29 &24.28 &19.17 &35.04\\
   IKL-AT &\textbf{63.40} &\textbf{55.31} &\textbf{46.76} &\textbf{38.98} &\textbf{31.91} &\textbf{25.33} &\textbf{19.98} &\textbf{36.38}\\
\bottomrule
\end{tabular}
}
\label{tab:ablation_epsilon}
\vspace{-0.1in}
\end{table}

\begin{table*}[tb!]
\centering
\caption{\textbf{Evaluation under PGD and CW attacks.}}
\vspace{-0.1in}
\resizebox{1.0\linewidth}{!}
{
\begin{tabular}{cccccccc}
\toprule
 Method &Acc &PGD-10 &PGD-20 &CW-10 &CW-20 &Auto-Attack &Worst \\
\midrule
KL(TRADES)  &62.87 &36.01 &35.84 &40.03 &39.86 &30.29 &30.29\\
\midrule
IKL(Ours)                     &63.40 &36.78  &36.55 &40.72 &40.47 &31.91 &31.92\\
IKL(Ours with autoaug)        &65.93 &38.15  &37.75 &41.10 &40.86 &32.53 &32.52\\
IKL(Ours with synthetic data) &73.85 &44.43  &44.12 &47.59 &47.53 &39.18 &39.18 \\
\bottomrule
\end{tabular}
}
\label{tab:pgd_cw}
\vspace{-0.2in}
\end{table*}

{\bf Robustness under Other Attacks.}
Auto-Attack is currently one of the strongest attack methods. It ensembles several adversarial attack methods including APGD-CE, APGD-DLR, FAB, and Square Attack. To show the effectiveness of our IKL loss, we also evaluate our models under PGD and CW attacks with 10 and 20 iterations. The perturbation size and step size are set to 8/255 and 2/255 respectively. As shown in Table~\ref{tab:pgd_cw}, with increasing iterations from 10 to 20, our models show similar robustness, demonstrating that our models don't suffer from obfuscated gradients problem. 
\vspace{-0.05in}
\section{Conclusion and Limitation}
\label{sec:conclusion}
\vspace{-0.05in}
In this paper, we have investigated the mechanism of Kullback-Leibler (KL) Divergence loss in terms of gradient optimization.
Based on our analysis, we decouple the KL loss into a weighted Mean Square Error ($\mathbf{w}$MSE) loss and a Cross-Entropy loss with soft labels.
The new formulation is named Decoupled Kullback-Leibler (DKL) Divergence loss.
To address the spotted issues of KL/DKL, we make two improvements that break the asymmetric optimization property and incorporate class-wise global information,
deriving the Improved Kullback-Leibler (IKL) Divergence loss.
Experimental results on CIFAR-10/100 and ImageNet show that we create new state-of-the-art adversarial robustness and competitive performance on knowledge distillation.
This underscores the efficacy of our Innovative KL (IKL) loss technique.
The KL loss exhibits a wide range of applications.
As part of our future work, we aim to explore and highlight the versatility of IKL in various other scenarios, like robustness on out-of-distribution data,  and incremental learning.

{
\small
\bibliography{egbib}
\bibliographystyle{unsrt}
}

%%%%%%%%%%%%%%%%%%%%%%%%%%%%%%%%%%%%%%%%%%%%%%%%%%%%%%%%%%%%
\newpage
\appendix

\section{Appendix}

\subsection{Proof to Theorem~\ref{thm:thm_dkl}}
\label{sec:proof}
To demonstrate that DKL in Eq.~\eqref{eq_dkl} is equivalent to KL in Eq.~\eqref{eq_kl_sample} for training optimization, we prove that DKL and KL produce the same gradients when given the same inputs.

For KL loss, we have the following derivatives according to the chain rule:
\begin{eqnarray}
    \frac{\partial \mathbf{s}_{m}^{i}}{\partial \mathbf{o}_{m}^{i}} &=& \mathbf{s}_{m}^{i} * \sum_{j!=i}^{C} \mathbf{s}_{m}^{j},  \nonumber \\
    \frac{\partial \mathbf{s}_{m}^{j}}{\partial \mathbf{o}_{m}^{i}} &=& -\mathbf{s}_{m}^{i} * \mathbf{s}_{m}^{j}, \nonumber \\
    \frac{\partial \mathcal{L}_{KL}}{\partial \mathbf{s}_{m}^{i}} &=& \log \mathbf{s}_{m}^{i} - \log \mathbf{s}_{n}^{i} +1, \nonumber \\
    \frac{\partial \mathcal{L}_{KL}}{\partial \mathbf{o}_{n}^{i}} &=& \mathbf{s}_{m}^{i} * (\mathbf{s}_{n}^{i} -1) + \mathbf{s}_{n}^{i} * (1 - \mathbf{s}_{m}^{i}) \label{kl_n} \\
    \frac{\partial \mathcal{L}_{KL}}{\partial \mathbf{o}_{m}^{i}} &=& \frac{\mathcal{L}_{KL}}{\partial \mathbf{s}_{m}^{i}} * \frac{\partial \mathbf{s}_{m}^{i}}{\partial \mathbf{o}_{m}^{i}} + \sum_{j!=i}^{C} \frac{\mathcal{L}_{KL}}{\partial \mathbf{s}_{m}^{j}} * \frac{\partial \mathbf{s}_{m}^{j}}{\partial \mathbf{o}_{m}^{i}} \nonumber \\
    &=& (\log \mathbf{s}_{m}^{i} - \log \mathbf{s}_{n}^{i} + 1) * \mathbf{s}_{m}^{i} * \sum_{j!=i}^{C} \mathbf{s}_{m}^{j} + \sum_{j!=i}^{C} (\log \mathbf{s}_{m}^{j} - \log \mathbf{s}_{n}^{j} +1) * (-\mathbf{s}_{m}^{j} * \mathbf{s}_{m}^{i}) \nonumber \\
    &=& \sum_{i!=j}^{C} ((\log \mathbf{s}_{m}^{i} - \log \mathbf{s}_{m}^{j}) - (\log \mathbf{s}_{n}^{i} - \log \mathbf{s}_{n}^{j})) * (\mathbf{s}_{m}^{j} * \mathbf{s}_{m}^{i}) \nonumber \\
    &=& \sum_{i!=j}^{C} ((\mathbf{o}_{m}^{i} - \mathbf{o}_{m}^{j}) - (\mathbf{o}_{n}^{i} - \mathbf{o}_{n}^{j})) * (\mathbf{s}_{m}^{j} * \mathbf{s}_{m}^{i}) \nonumber \\
    &=& \sum_{i!=j}^{C}(\Delta \mathbf{m}_{i,j} - \Delta \mathbf{n}_{i,j}) * \mathbf{w}_{m}^{i,j} \nonumber \\
    &=& \sum_{j}^{C}(\Delta \mathbf{m}_{i,j} - \Delta \mathbf{n}_{i,j}) * \mathbf{w}_{m}^{i,j}
    \label{kl_m}
\end{eqnarray}

For DKL los, we expand the Eq.~\eqref{eq_dkl} as:
\begin{eqnarray}
            \mathcal{L}_{DKL}(x_m, x_n)  &=& \underbrace{\frac{\alpha}{4} \|\sqrt{\mathcal{S}(\mathbf{w}_{m})}(\Delta \mathbf{m} -\mathcal{S}(\Delta \mathbf{n}))\|^2}_{\textbf{weighted MSE (wMSE)}}  
            \underbrace{-\beta \cdot \mathcal{S}(\mathbf{s}_{m}^{\top}) \cdot \log \mathbf{s}_{n}}_{\textbf{Cross-Entropy}} \\
            &=& \underbrace{\frac{\alpha}{4} \sum_{j=1}^{C} \sum_{k=1}^{C} ((\Delta \mathbf{m}_{j,k} \!-\! \mathcal{S}(\Delta \mathbf{n}_{j,k}) )^2 * \mathcal{S}(\mathbf{w}_{m}^{j,k}))}_{\textbf{weighted MSE (wMSE)}} \underbrace{-\beta \sum_{j=1}^{C} \mathcal{S}(\mathbf{s}_{m}^{j}) * \log \mathbf{s}_{n}^{j}}_{\textbf{Cross-Entropy }} \nonumber,
        \label{eq_dkl_expand}
\end{eqnarray}

According to the chain rule, we obtain the following equations:
\begin{eqnarray}
\frac{\partial \mathcal{L}_{DKL}}{\partial \mathbf{o}_{n}^{i}} &=& \beta * \mathbf{s}_{m}^{i} * (\mathbf{s}_{n}^{i} -1) + \mathbf{s}_{n}^{i} * (1 - \mathbf{s}_{m}^{i})  \label{dkl_n}\\
\frac{\partial \mathcal{L}_{DKL}}{\partial \mathbf{o}_{m}^{i}} &=& \frac{\alpha}{4} * 2 * (\sum_{j}^{C} (\Delta \mathbf{m}_{j,i}-\Delta \mathbf{n}_{j,i}) * (-\mathbf{w}_{m}^{j,i}) + \sum_{k}^{C} (\Delta \mathbf{m}_{i,k} - \Delta \mathbf{n}_{i,k}) * \mathbf{w}_{m}^{i,k})\nonumber \\
&=& \alpha * \sum_{j}^{C}(\Delta \mathbf{m}_{i,j} - \Delta \mathbf{n}_{i,j}) * \mathbf{w}_{m}^{i,j}
\label{dkl_m}
\end{eqnarray}

Comparing Eq.~\eqref{kl_n} and Eq.~\eqref{dkl_n}, Eq.~\eqref{kl_m} and Eq.~\eqref{dkl_m}, we conclude that
DKL loss and KL loss have the same derivatives given the same inputs. Thus, DKL loss is equivalent to KL loss in terms of gradient optimization.

\subsection{New state-of-the-art robustness on CIFAR-100/10}
\vspace{+0.1in}
\textit{Robustbench} is the most popular benchmark for adversarial robust models in the community. It evaluates the performance of models by the Auto-Attack. Auto-Attack \cite{croce2020reliable} is an ensemble of different kinds of attack methods and is considered the most effective method to test the robustness of models.

We achieve new state-of-the-art robustness on CIFAR-10 and CIFAR-100 under both settings w/ and w/o generated data.
As shown in Table~\ref{tab:comparision_sota}, on CIFAR-100 without extra generated data, we achieve 32.53\% robustness, outperforming the previous best result by \textbf{0.68\%} while saving \textbf{33.3\%} computational cost. With generated data, our model boosts performance to 73.85\% natural accuracy, surpassing the previous best result by \textbf{1.27\%} while maintaining the \textbf{strongest robustness}. More detailed comparisons can be accessed on the public leaderboard \url{https://robustbench.github.io/}.

\begin{table*}[tb!]
\centering
\caption{\textbf{New state-of-the-art on public leaderboard { \textit RobustBench} \cite{croce2020reliable}.}}
\label{tab:comparision_sota}
\resizebox{1.0\linewidth}{!}
{
\begin{tabular}{ccccc}
\toprule
 Experimental Settings & augmentation strategy & Clean  & AA &Computation saving\\
\midrule
    w/o Generated Data (Previous best results)        & Basic   &62.99  &31.20 &\\
    w/o Generated Data (Ours)                         & Basic   &\textbf{65.76(+2.67)}  &\textbf{31.91(+0.71)} &\textbf{33.3\%} \\
    \midrule
    w/o Generated Data (Previous best results)        & Autoaug   &\textbf{68.75}  &31.85 &\\
    w/o Generated Data (Ours)                         & Autoaug   &66.08  &\textbf{32.53(+0.68)} &\textbf{33.3\%}\\
    \midrule
    w/ Generated Data (Previous best results)    & Genreated data &72.58  &38.83 &\\ 
    w/ Generated Data (Ours)                     & Generated data &\textbf{73.85(+1.27)}  &\textbf{39.18(+0.35)} &0\%\\ 
\bottomrule
\end{tabular}
}
\end{table*}

\begin{table*}[tb!]
%\setlength{\belowcaptionskip}{-10pt} %endtabspace
%\small
\center
\caption{\textbf{Top-1 accuracy~(\%) on the CIFAR-100 validation.} Teachers and students are in the \textbf{same} architectures. All results are the average over three trials.}
\resizebox{.95\linewidth}{!}
{
\begin{tabular}{cccccccc}
\toprule
\multirow{4}{*}{\begin{tabular}[c]{@{}c@{}}Distillation \\ Manner\end{tabular}} & \multirow{2}{*}{Teacher}  & ResNet56 & ResNet110 & ResNet32$\times$4 & WRN-40-2 & WRN-40-2 & VGG13 \\
&      & 72.34      & 74.31       & 79.42        & 75.61      & 75.61      & 74.64   \\
& \multirow{2}{*}{Student}  & ResNet20 & ResNet32 & ResNet8$\times$4  & WRN-16-2 & WRN-40-1 & VGG8  \\
& \space     & 69.06      & 71.14       & 72.50        & 73.26      & 71.98      & 70.36   \\
\midrule
\multirow{5}{*}{Features}
& FitNet   & 69.21      & 71.06       & 73.50        & 73.58      & 72.24      & 71.02   \\
& RKD          & 69.61      & 71.82       & 71.90        & 73.35      & 72.22      & 71.48   \\
& CRD          & 71.16      & 73.48       & 75.51        & 75.48      & 74.14      & 73.94   \\
& OFD          & 70.98      & 73.23       & 74.95        & 75.24      & 74.33      & 73.95   \\
& ReviewKD   & 71.89      & 73.89       & 75.63        & 76.12      & \textbf{75.09}      & 74.84   \\ 
\midrule
\multirow{4}{*}{Logits}                                                         
& DKD        & \textbf{71.97}      & 74.11       & 76.32              & 76.24          & 74.81      & 74.68 \\
& KD          & 70.66               & 73.08                & 73.33              & 74.92          & 73.54      & 72.98   \\
& \textbf{IKL-KD}      & 71.44               & \textbf{74.26}                & \textbf{76.59}     & \textbf{76.45} & 74.98      & \textbf{74.98} \\
%& $\Delta$             &\textcolor{myorange}{+0.78}  &\textcolor{myorange}{+1.16}  &\textcolor{myorange}{+3.26} &\textcolor{myorange}{+1.53} &\textcolor{myorange}{+1.44} &\textcolor{myorange}{+2.00} \\
\bottomrule
\end{tabular}
}
\label{tab:cifar_kd}
\end{table*}

\begin{table*}[tb!]
\center
%\small
\caption{\textbf{Top-1 accuracy~(\%) on the CIFAR-100 validation.} Teachers and students are in \textbf{different} architectures. All results are the average over 3 trials.}
\resizebox{1.0\linewidth}{!}
{
\begin{tabular}{ccccccc}
\toprule
\multirow{4}{*}{\begin{tabular}[c]{@{}c@{}}Distillation \\ Manner\end{tabular}} & \multirow{2}{*}{Teacher}  & ResNet32$\times$4 & WRN-40-2 & VGG13 & ResNet50 & ResNet32$\times$4 \\
& \space     & 79.42      & 75.61       & 74.64        & 79.34      & 79.42         \\
& \multirow{2}{*}{Student}  & ShuffleNet-V1 & ShuffleNet-V1 & MobileNet-V2  & MobileNet-V2 & ShuffleNet-V2  \\
& \space     & 70.50      & 70.50       & 64.60        & 64.60      & 71.82         \\ 
\midrule
\multirow{5}{*}{Features}
& FitNet & 73.59      & 73.73       & 64.14        & 63.16      & 73.54    \\
& RKD        & 72.28      & 72.21       & 64.52        & 64.43      & 73.21    \\
& CRD        & 75.11      & 76.05       & 69.73        & 69.11      & 75.65    \\
& OFD        & 75.98      & 75.85       & 69.48        & 69.04      & 76.82    \\
& ReviewKD & \textbf{77.45}      & 77.14       & 70.37        & 69.89      & \textbf{77.78}    \\     
\midrule
\multirow{4}{*}{Logits}                                                         
& DKD     & 76.45               & 76.70                        & 69.71          & 70.35              & 77.07       \\
& KD        & 74.07               & 74.83                        & 67.37          & 67.35              & 74.45         \\
& \textbf{IKL-KD}    & 76.64 $\pm$ 0.02      & \textbf{77.19} $\pm$ 0.01      & \textbf{70.40} $\pm$ 0.03    & \textbf{70.62} $\pm$ 0.08   & 77.16 $\pm$ 0.04 \\
%& $\Delta$           & \textcolor{myorange}{+2.57} &\textcolor{myorange}{+2.63} &\textcolor{myorange}{+3.03} &\textcolor{myorange}{+3.27} &\textcolor{myorange}{+2.71} \\
\bottomrule
\end{tabular}
}
\label{tab:cifar2_kd}
%\vspace{+0.2in}
\end{table*}

\begin{table*}[h]
\centering
\small
\caption{\textbf{Comparisons with strong training settings on ImageNet for knowledge distillation.}}
\label{tab:comparision_strong}
%\resizebox{1.0\linewidth}{!}
{
\begin{tabular}{ccccc}
\toprule
 Method & KD & DKD &DIST &IKL-KD \\
\midrule
  Top-1 Accuracy (\%) & 80.89 & 80.77 &80.70 & \textbf{80.98} \\
\bottomrule
\end{tabular}
}
%\vspace{+0.2in}
\end{table*}

\subsection{Comparisons on CIFAR-100 for Knowledge Distillation}
We experiment on CIFAR-100 with the following cases: 1) the teacher and student models have the same unit network architectures; 2) the teacher and student models have different unit network architectures. The results are listed in Table~\ref{tab:cifar_kd} and Table~\ref{tab:cifar2_kd}. We have achieved the best results in 4 out of 6 and 3 out of 5 experimental settings respectively.

Moreover, we follow the concurrent work~\cite{hao2023vanillakd} and conduct experiments with BEiT-Large as the teacher and ResNet-50 as the student under a strong training scheme, the experimental results are summarized in Table~\ref{tab:comparision_strong}. The model trained by IKL-KD shows slightly better results.

\subsection{Other Applications with IKL}
\iffalse
\noindent{\bf Knowledge Distillation on Imabalanced Data.} 
For the long-tailed recognition on ImageNet-LT, we train models 90 epochs with cross-entropy loss. We only preprocess images with \textit{RandomResizedCrop} and random horizontal flip.  All training settings are the same for fair comparisons. As shown in Table~\ref{tab:dkl_imagenetlt}, Models trained with our loss function achieve much better performance than the original KL loss.

\begin{table*}[h]
\centering
\caption{\textbf{ Experiments on Long-tailed Data (ImageNet-LT)}.}
%\resizebox{1.0\linewidth}{!}
{
\begin{tabular}{ccccc}
\toprule
 Method &Student   &Teacher   &Teacher Acc(\%)  & Student Acc(\%) \\
\midrule
\multicolumn{5}{c}{\textbf{Self-distillation on Imbalanced Data}} \\
\midrule
 KD with KL      &ResNet-50 &ResNet-50  &45.47  &46.50 \\
 KD with JSD     &ResNet-50 &ResNet-50  &45.47  &45.98 \\
 KD with our IKL &ResNet-50 &ResNet-50  &45.47  &\textbf{47.46(+0.96)} \\
 \midrule
 \multicolumn{5}{c}{\textbf{Knowledge distillation on Imbalanced Data}} \\
 \midrule
 KD with KL            &ResNet-50 &ResNeXt-101 &48.33 & 47.10 \\
 KD with JSD           &ResNet-50 &ResNeXt-101 &48.33 & 46.51 \\
 KD with our IKL       &ResNet-50 &ResNeXt-101 &48.33 & \textbf{48.54(+1.44)} \\
\bottomrule
\end{tabular}
}
\label{tab:dkl_imagenetlt}
\end{table*}
\fi

\noindent{\bf Semisupervised learning.}
We use the open-sourced code from \url{https://github.com/microsoft/Semi-supervised-learning} and conduct semi-supervised experiments on CIFAR-100 with FixMatch and Mean-Teacher methods. Specifically, each class has 2 labeled images and 500 unlabeled images. All default training hyper-parameters are used for fair comparisons. We only replace the consistency loss with our IKL loss. As shown in Table~\ref{tab:dkl_semi}, with our IKL loss, the Mean-Teacher method even surpasses the FixMatch.

\begin{table*}[h]
\centering
\caption{\textbf{Semi-supervised Learning on CIFAR-100 with ViT-small backbone.}}
%\resizebox{1.0\linewidth}{!}
{
\begin{tabular}{cccc}
\toprule
 Method  & Pseudo-label &Consistency Loss   &Last epoch Top-1 Acc(\%) \\
\midrule
\multicolumn{4}{c}{\textbf{FixMatch}} \\
\midrule
FixMatch       &hard  &Cross-entropy Loss    &69.20 \\
\midrule
FixMatch       &soft  &Cross-entroy/KL Loss  &69.09 \\
FixMatch       &soft  &IKL Loss              &\textbf{70.00} \\
 \midrule
 \multicolumn{4}{c}{\textbf{Mean-Teacher}} \\
 \midrule
Mean-Teacher  &soft  &MSE Loss               &67.38 \\
Mean-Teacher  &soft  &IKL Loss               &\textbf{70.05}      \\
\bottomrule
\end{tabular}
}
\label{tab:dkl_semi}
\end{table*}

\noindent{\bf Semantic segmentation distillation.} We conduct ablation on the semantic segmentation distillation task. We use the APD~\cite{tian2022adaptive} as our baseline for their open-sourced code. All default hyper-parameters are adopted. We only replace the original KL loss with our IKL loss. As shown in Table~\ref{tab:dkl_seg}, we achieve better performance with the IKL loss function, demonstrating that the IKL loss can be complementary to other techniques in semantic segmentation distillation.

\begin{table*}[h]
\centering
\caption{\textbf{Semantic segmentation distillation with APD on ADE20K.}}
%\resizebox{1.0\linewidth}{!}
{
\begin{tabular}{ccccc}
\toprule
 Method  &Teacher &Student & Teacher mIoU   &Student mIoU \\
\midrule
Baseline                 &-          &ResNet-18 &-     &37.19  \\              
\midrule
APD with KL loss  &ResNet-101 &ResNet-18 &43.44 &39.25 \\
APD with IKL loss &ResNet-101 &ResNet-18 &43.44 &\textbf{39.75} \\
\bottomrule
\end{tabular}
}
\label{tab:dkl_seg}
\end{table*}

\subsection{Complexity of IKL}
Compared with the KL divergence loss, IKL loss is required to update the global class-wise prediction scores $W \in \mathbb{R}^{C \times C}$ where $C$ is the number of classes during training. 
This extra computational cost can be nearly ignored when compared with the model forward and backward. Algorithm~\ref{algo:algorithm_dkl} shows the implementation of our IKL loss in Pytorch style. On dense prediction tasks like semantic segmentation, $\Delta_{a}$ and $\Delta_{b}$ can require large GPU Memory. Here, we also provide the memory-efficient implementations for $w$MSE loss component, which is listed in Algorithm~\ref{algo:algorithm_wmse}.

\begin{algorithm}
    \caption{Pseudo code for DKL/IKL loss in Pytorch style.}
    \begin{algorithmic}
        \State \textbf{Input:} $logits_{a}, logits_{b} \in \mathbb{R}^{B \times C}$, one-hot label $Y$, $W \in \mathbb{R}^{C \times C}$, $\alpha$, $\beta$. 

        \State class\_scores = one-hot @ W; 
        \State Sample\_weights = class\_scores.view(-1, C, 1) @ class\_scores.view(-1, 1, C); 
        \State $\Delta_a$ = $logits_a$.view(-1, C, 1) - $logits_a$.view(-1, 1, C);
        \State $\Delta_b$ = $logits_b$.view(-1, C, 1) - $logits_b$.view(-1, 1, C);
        \State wMSE\_loss = (torch.pow($\Delta_{n}$ - $\Delta_{a}$, 2) * Sample\_weights).sum(dim=(1,2)).mean() * $\frac{1}{4}$;
        
        \State score\_a = F.softmax($logits_a$, dim=1).detach();
        \State log\_score\_b = F.log\_softmax($logits_b$, dim=-1);
        \State CE\_loss = -(score\_a * log\_score\_b).sum(1).mean(); 
        \State $\textbf{return}$ $\beta$ * CE\_loss + $\alpha$ * wMSE\_loss.
    \end{algorithmic}
\label{algo:algorithm_dkl}
\end{algorithm}

\begin{algorithm}
    \caption{Memory efficient implementation for wMSE\_loss in Pytorch style.}
    \begin{algorithmic}
        \State \textbf{Input:} $logits_{a}, logits_{b} \in \mathbb{R}^{B \times C}$, one-hot label $Y$, $W \in \mathbb{R}^{C \times C}$;

        \State class\_scores = one-hot @ W; 
        \State loss\_a = (class\_scores * $logits_a$ * $logits_a$).sum(dim=1) * 2 - torch.pow((class\_scores * $logits_a$).sum(dim=1), 2) * 2;
        \State loss\_b = (class\_scores * $logits_b$ * $logits_b$).sum(dim=1) * 2 - torch.pow((class\_scores * $logits_b$).sum(dim=1), 2) * 2;
        \State loss\_ex = (class\_scores * $logits_a$ * $logits_b$).sum(dim=1) * 4 - (class\_scores * $logits_a$).sum(dim=1) * (class\_scores * $logits_b$).sum(dim=1) * 4;
        \State wMSE\_loss = $\frac{1}{4}$ * (loss\_a + loss\_b - loss\_ex).mean();
        \State $\textbf{return}$ wMSE\_loss.
    \end{algorithmic}
    \label{algo:algorithm_wmse}
\end{algorithm}

\subsection{\bf Connection between IKL and the Jensen-Shannon (JS) Divergence}
With the following JS divergence loss,
\begin{equation}
    JSD(P||Q) = \frac{1}{2} KL(P||M) + \frac{1}{2} KL(Q||M), \quad M=\frac{1}{2} P + \frac{1}{2} Q.
\end{equation}

We calculate its derivatives regarding $o_{n}$ (the student logits), 
\begin{eqnarray}
  \frac{\partial \mathcal{L}_{JSD}}{\partial \mathbf{o}_{n}^{i}} &=& \sum_{j=1}^{C}\mathbf{w}_{n}^{i,j} (\Delta \mathbf{n}_{i,j} - \Delta \mathbf{m'}_{i,j}) \label{eq:jsd_gradient}\\
  \textit{Softmax}(o_{m'}) &=& \frac{1}{2} s_{n} + \frac{1}{2} s_{m} \label{eq:jsd_constraint}
\end{eqnarray}
where $\mathbf{o}_{m}$ is the logits from the teacher model, $\mathbf{o}_{m'}$ is a virtual logits satisfying Eq.~\eqref{eq:jsd_constraint}, $\mathbf{s}_{m}=\textit{Softmax}(\mathbf{o}_{m})$, $\mathbf{s}_{n}=\textit{Softmax}(\mathbf{o}_{n})$, $\Delta \mathbf{m'}_{i,j}= \mathbf{o}_{m'}^{i} - \mathbf{o}_{m'}^{j}$, $\Delta \mathbf{n}_{i,j}= \mathbf{o}_{n}^{i} - \mathbf{o}_{n}^{j}$.

Correspondingly, the derivatives of IKL loss regrading $o_{n}$ (the student logits),
\begin{eqnarray}
    \frac{\partial \mathcal{L}_{IKL}}{\partial \mathbf{o}_{n}^{i}} = \underbrace{\alpha \sum_{j=1}^{C} \mathbf{w}_{m}^{i,j}(\Delta \mathbf{n}_{i,j} - \Delta \mathbf{m}_{i,j})}_{\textbf{Effects of wMSE}} + \underbrace{\beta * \mathbf{s}_{m}^{i} * (\mathbf{s}_{n}^{i} -1) + \mathbf{s}_{n}^{i} * (1-\mathbf{s}_{m}^{i})}_{\textbf{Effects of Cross-Entropy}}
\end{eqnarray}

Compared with IKL loss, the problem for JSD divergence in knowledge distillation is that:
\textit{The soft labels from the teacher models often embed dark knowledge and facilitate the optimization of the student models. However, there are no effects of the cross-entropy loss with the soft labels from the teacher model, which can be the underlying reason that JSD is worse than KD and IKL-KD.}

As shown in Table~\ref{tab:jsd_vs_kl_dkl}, we also empirically demonstrate that IKL loss performs better than JSD divergence on the knowledge distillation task.

\begin{table*}[h]
\centering
\caption{\textbf{Comparisons between KL, IKL, and JSD on ImageNet-LT}.}
%\resizebox{1.0\linewidth}{!}
{
\begin{tabular}{ccccc}
\toprule
 Method &Student   &Teacher   &Teacher Acc(\%)  & Student Acc(\%) \\
\midrule
\multicolumn{5}{c}{\textbf{Self-distillation on Imbalanced Data}} \\
\midrule
 KL   &ResNet-50 &ResNet-50  &45.47  &47.04 \\
 JSD  &ResNet-50 &ResNet-50  &45.47  &46.64 \\
 Ours &ResNet-50 &ResNet-50  &45.47  &\textbf{47.50} \\
 \midrule
 \multicolumn{5}{c}{\textbf{Knowledge distillation on Imbalanced Data}} \\
 \midrule
 KL   &ResNet-50 &ResNeXt-101 &48.33 &48.31 \\
 JSD  &ResNet-50 &ResNeXt-101 &48.33 &47.82 \\
 Ours &ResNet-50 &ResNeXt-101 &48.33 & \textbf{49.22} \\
\bottomrule
\end{tabular}
}
\label{tab:jsd_vs_kl_dkl}
\end{table*}

\subsection{Licenses}\label{app:licenses}
All the datasets we considered are publicly available, we list their licenses and URLs as follows:

\begin{itemize}
    \item \textbf{CIFAR-10}~\cite{krizhevsky2009learning}: MIT License, \url{https://www.cs.toronto.edu/~kriz/cifar.html}.
    \item \textbf{CIFAR-100}~\cite{krizhevsky2009learning}: MIT License,
    \url{https://www.cs.toronto.edu/~kriz/cifar.html}.
    \item \textbf{ImageNet}~\cite{imagenet}: Non-commercial, \url{http://image-net.org}.
   
\end{itemize}

%Optionally include supplemental material (complete proofs, additional experiments and plots) in appendix.
%All such materials \textbf{SHOULD be included in the main submission.}

%%%%%%%%%%%%%%%%%%%%%%%%%%%%%%%%%%%%%%%%%%%%%%%%%%%%%%%%%%%%
\section*{NeurIPS Paper Checklist}

\begin{enumerate}

\item {\bf Claims}
    \item[] Question: Do the main claims made in the abstract and introduction accurately reflect the paper's contributions and scope?
    \item[] Answer: \answerYes{} % Replace by \answerYes{}, \answerNo{}, or \answerNA{}.
    \item[] Justification: We theoretically prove that the Kullback-Leibler (KL) Divergence loss is equivalent to a Decoupled Kullback-Leibler (DKL) loss regarding gradient optimization. Based on this analysis, we improve KL/DKL loss on adversarial training and knowledge distillation tasks. %\justificationTODO{}
    \item[] Guidelines:
    \begin{itemize}
        \item The answer NA means that the abstract and introduction do not include the claims made in the paper.
        \item The abstract and/or introduction should clearly state the claims made, including the contributions made in the paper and important assumptions and limitations. A No or NA answer to this question will not be perceived well by the reviewers. 
        \item The claims made should match theoretical and experimental results, and reflect how much the results can be expected to generalize to other settings. 
        \item It is fine to include aspirational goals as motivation as long as it is clear that these goals are not attained by the paper. 
    \end{itemize}

\item {\bf Limitations}
    \item[] Question: Does the paper discuss the limitations of the work performed by the authors?
    \item[] Answer: \answerYes{} % Replace by \answerYes{}, \answerNo{}, or \answerNA{}.
    \item[] Justification: The conclusion and limitation section is included.%\justificationTODO{}
    \item[] Guidelines:
    \begin{itemize}
        \item The answer NA means that the paper has no limitation while the answer No means that the paper has limitations, but those are not discussed in the paper. 
        \item The authors are encouraged to create a separate "Limitations" section in their paper.
        \item The paper should point out any strong assumptions and how robust the results are to violations of these assumptions (e.g., independence assumptions, noiseless settings, model well-specification, asymptotic approximations only holding locally). The authors should reflect on how these assumptions might be violated in practice and what the implications would be.
        \item The authors should reflect on the scope of the claims made, e.g., if the approach was only tested on a few datasets or with a few runs. In general, empirical results often depend on implicit assumptions, which should be articulated.
        \item The authors should reflect on the factors that influence the performance of the approach. For example, a facial recognition algorithm may perform poorly when image resolution is low or images are taken in low lighting. Or a speech-to-text system might not be used reliably to provide closed captions for online lectures because it fails to handle technical jargon.
        \item The authors should discuss the computational efficiency of the proposed algorithms and how they scale with dataset size.
        \item If applicable, the authors should discuss possible limitations of their approach to address problems of privacy and fairness.
        \item While the authors might fear that complete honesty about limitations might be used by reviewers as grounds for rejection, a worse outcome might be that reviewers discover limitations that aren't acknowledged in the paper. The authors should use their best judgment and recognize that individual actions in favor of transparency play an important role in developing norms that preserve the integrity of the community. Reviewers will be specifically instructed to not penalize honesty concerning limitations.
    \end{itemize}

\item {\bf Theory Assumptions and Proofs}
    \item[] Question: For each theoretical result, does the paper provide the full set of assumptions and a complete (and correct) proof?
    \item[] Answer: \answerYes{} % Replace by \answerYes{}, \answerNo{}, or \answerNA{}.
    \item[] Justification: Our proof to Theorem~\ref{thm:thm_dkl} is in Appendix~\ref{sec:proof}. %\justificationTODO{}
    \item[] Guidelines:
    \begin{itemize}
        \item The answer NA means that the paper does not include theoretical results. 
        \item All the theorems, formulas, and proofs in the paper should be numbered and cross-referenced.
        \item All assumptions should be clearly stated or referenced in the statement of any theorems.
        \item The proofs can either appear in the main paper or the supplemental material, but if they appear in the supplemental material, the authors are encouraged to provide a short proof sketch to provide intuition. 
        \item Inversely, any informal proof provided in the core of the paper should be complemented by formal proofs provided in appendix or supplemental material.
        \item Theorems and Lemmas that the proof relies upon should be properly referenced. 
    \end{itemize}

    \item {\bf Experimental Result Reproducibility}
    \item[] Question: Does the paper fully disclose all the information needed to reproduce the main experimental results of the paper to the extent that it affects the main claims and/or conclusions of the paper (regardless of whether the code and data are provided or not)?
    \item[] Answer: \answerYes{} % Replace by \answerYes{}, \answerNo{}, or \answerNA{}.
    \item[] Justification: We provide detailed experimental settings in Section~\ref{sec:exp}. %\justificationTODO{}
    \item[] Guidelines:
    \begin{itemize}
        \item The answer NA means that the paper does not include experiments.
        \item If the paper includes experiments, a No answer to this question will not be perceived well by the reviewers: Making the paper reproducible is important, regardless of whether the code and data are provided or not.
        \item If the contribution is a dataset and/or model, the authors should describe the steps taken to make their results reproducible or verifiable. 
        \item Depending on the contribution, reproducibility can be accomplished in various ways. For example, if the contribution is a novel architecture, describing the architecture fully might suffice, or if the contribution is a specific model and empirical evaluation, it may be necessary to either make it possible for others to replicate the model with the same dataset, or provide access to the model. In general. releasing code and data is often one good way to accomplish this, but reproducibility can also be provided via detailed instructions for how to replicate the results, access to a hosted model (e.g., in the case of a large language model), releasing of a model checkpoint, or other means that are appropriate to the research performed.
        \item While NeurIPS does not require releasing code, the conference does require all submissions to provide some reasonable avenue for reproducibility, which may depend on the nature of the contribution. For example
        \begin{enumerate}
            \item If the contribution is primarily a new algorithm, the paper should make it clear how to reproduce that algorithm.
            \item If the contribution is primarily a new model architecture, the paper should describe the architecture clearly and fully.
            \item If the contribution is a new model (e.g., a large language model), then there should either be a way to access this model for reproducing the results or a way to reproduce the model (e.g., with an open-source dataset or instructions for how to construct the dataset).
            \item We recognize that reproducibility may be tricky in some cases, in which case authors are welcome to describe the particular way they provide for reproducibility. In the case of closed-source models, it may be that access to the model is limited in some way (e.g., to registered users), but it should be possible for other researchers to have some path to reproducing or verifying the results.
        \end{enumerate}
    \end{itemize}

\item {\bf Open access to data and code}
    \item[] Question: Does the paper provide open access to the data and code, with sufficient instructions to faithfully reproduce the main experimental results, as described in supplemental material?
    \item[] Answer: \answerYes{} % Replace by \answerYes{}, \answerNo{}, or \answerNA{}.
    \item[] Justification: We provide a link for our code in the paper. %\justificationTODO{}
    \item[] Guidelines:
    \begin{itemize}
        \item The answer NA means that paper does not include experiments requiring code.
        \item Please see the NeurIPS code and data submission guidelines (\url{https://nips.cc/public/guides/CodeSubmissionPolicy}) for more details.
        \item While we encourage the release of code and data, we understand that this might not be possible, so “No” is an acceptable answer. Papers cannot be rejected simply for not including code, unless this is central to the contribution (e.g., for a new open-source benchmark).
        \item The instructions should contain the exact command and environment needed to run to reproduce the results. See the NeurIPS code and data submission guidelines (\url{https://nips.cc/public/guides/CodeSubmissionPolicy}) for more details.
        \item The authors should provide instructions on data access and preparation, including how to access the raw data, preprocessed data, intermediate data, and generated data, etc.
        \item The authors should provide scripts to reproduce all experimental results for the new proposed method and baselines. If only a subset of experiments are reproducible, they should state which ones are omitted from the script and why.
        \item At submission time, to preserve anonymity, the authors should release anonymized versions (if applicable).
        \item Providing as much information as possible in supplemental material (appended to the paper) is recommended, but including URLs to data and code is permitted.
    \end{itemize}

\item {\bf Experimental Setting/Details}
    \item[] Question: Does the paper specify all the training and test details (e.g., data splits, hyperparameters, how they were chosen, type of optimizer, etc.) necessary to understand the results?
    \item[] Answer: \answerYes{} % Replace by \answerYes{}, \answerNo{}, or \answerNA{}.
    \item[] Justification: We provide detailed experimental settings in Section~\ref{sec:exp} and ablations for hyper-parameters in the Appendix~\ref{sec:ablation}.%\justificationTODO{}
    \item[] Guidelines:
    \begin{itemize}
        \item The answer NA means that the paper does not include experiments.
        \item The experimental setting should be presented in the core of the paper to a level of detail that is necessary to appreciate the results and make sense of them.
        \item The full details can be provided either with the code, in appendix, or as supplemental material.
    \end{itemize}

\item {\bf Experiment Statistical Significance}
    \item[] Question: Does the paper report error bars suitably and correctly defined or other appropriate information about the statistical significance of the experiments?
    \item[] Answer: \answerYes{} % Replace by \answerYes{}, \answerNo{}, or \answerNA{}.
    \item[] Justification: Error bars are included in Table~\ref{tab:cifar2_kd} of the Appendix. Due to the heavy computation, other tables don't provide error bars.%\justificationTODO{}
    \item[] Guidelines:
    \begin{itemize}
        \item The answer NA means that the paper does not include experiments.
        \item The authors should answer "Yes" if the results are accompanied by error bars, confidence intervals, or statistical significance tests, at least for the experiments that support the main claims of the paper.
        \item The factors of variability that the error bars are capturing should be clearly stated (for example, train/test split, initialization, random drawing of some parameter, or overall run with given experimental conditions).
        \item The method for calculating the error bars should be explained (closed form formula, call to a library function, bootstrap, etc.)
        \item The assumptions made should be given (e.g., Normally distributed errors).
        \item It should be clear whether the error bar is the standard deviation or the standard error of the mean.
        \item It is OK to report 1-sigma error bars, but one should state it. The authors should preferably report a 2-sigma error bar than state that they have a 96\% CI, if the hypothesis of Normality of errors is not verified.
        \item For asymmetric distributions, the authors should be careful not to show in tables or figures symmetric error bars that would yield results that are out of range (e.g. negative error rates).
        \item If error bars are reported in tables or plots, The authors should explain in the text how they were calculated and reference the corresponding figures or tables in the text.
    \end{itemize}

\item {\bf Experiments Compute Resources}
    \item[] Question: For each experiment, does the paper provide sufficient information on the computer resources (type of compute workers, memory, time of execution) needed to reproduce the experiments?
    \item[] Answer: \answerYes{} % Replace by \answerYes{}, \answerNo{}, or \answerNA{}.
    \item[] Justification: On adversarial training, Each run with basic augmentations takes around 2 days using 4GPUs while 5 days using 8 GPUs for adversarial training with generated data. On knowledge distillation, 8 Nvidia GeForce 3090 GPUs are used on ImageNet. Each run takes about 1 day for our method. %\justificationTODO{}
    \item[] Guidelines:
    \begin{itemize}
        \item The answer NA means that the paper does not include experiments.
        \item The paper should indicate the type of compute workers CPU or GPU, internal cluster, or cloud provider, including relevant memory and storage.
        \item The paper should provide the amount of compute required for each of the individual experimental runs as well as estimate the total compute. 
        \item The paper should disclose whether the full research project required more compute than the experiments reported in the paper (e.g., preliminary or failed experiments that didn't make it into the paper). 
    \end{itemize}
    
\item {\bf Code Of Ethics}
    \item[] Question: Does the research conducted in the paper conform, in every respect, with the NeurIPS Code of Ethics \url{https://neurips.cc/public/EthicsGuidelines}?
    \item[] Answer: \answerYes{} % Replace by \answerYes{}, \answerNo{}, or \answerNA{}.
    \item[] Justification: Our research conforms NeurIPS Code of Ethics.%\justificationTODO{}
    \item[] Guidelines:
    \begin{itemize}
        \item The answer NA means that the authors have not reviewed the NeurIPS Code of Ethics.
        \item If the authors answer No, they should explain the special circumstances that require a deviation from the Code of Ethics.
        \item The authors should make sure to preserve anonymity (e.g., if there is a special consideration due to laws or regulations in their jurisdiction).
    \end{itemize}

\item {\bf Broader Impacts}
    \item[] Question: Does the paper discuss both potential positive societal impacts and negative societal impacts of the work performed?
    \item[] Answer: \answerNA{} % Replace by \answerYes{}, \answerNo{}, or \answerNA{}.
    \item[] Justification: Our paper is fundamental research in adversarial robustness and knowledge distillation and there is no obvious societal impact.%\justificationTODO{}
    \item[] Guidelines:
    \begin{itemize}
        \item The answer NA means that there is no societal impact of the work performed.
        \item If the authors answer NA or No, they should explain why their work has no societal impact or why the paper does not address societal impact.
        \item Examples of negative societal impacts include potential malicious or unintended uses (e.g., disinformation, generating fake profiles, surveillance), fairness considerations (e.g., deployment of technologies that could make decisions that unfairly impact specific groups), privacy considerations, and security considerations.
        \item The conference expects that many papers will be foundational research and not tied to particular applications, let alone deployments. However, if there is a direct path to any negative applications, the authors should point it out. For example, it is legitimate to point out that an improvement in the quality of generative models could be used to generate deepfakes for disinformation. On the other hand, it is not needed to point out that a generic algorithm for optimizing neural networks could enable people to train models that generate Deepfakes faster.
        \item The authors should consider possible harms that could arise when the technology is being used as intended and functioning correctly, harms that could arise when the technology is being used as intended but gives incorrect results, and harms following from (intentional or unintentional) misuse of the technology.
        \item If there are negative societal impacts, the authors could also discuss possible mitigation strategies (e.g., gated release of models, providing defenses in addition to attacks, mechanisms for monitoring misuse, mechanisms to monitor how a system learns from feedback over time, improving the efficiency and accessibility of ML).
    \end{itemize}
    
\item {\bf Safeguards}
    \item[] Question: Does the paper describe safeguards that have been put in place for responsible release of data or models that have a high risk for misuse (e.g., pretrained language models, image generators, or scraped datasets)?
    \item[] Answer: \answerNA{} % Replace by \answerYes{}, \answerNo{}, or \answerNA{}.
    \item[] Justification: Our models don't suffer from a high risk for misuse. %\justificationTODO{}
    \item[] Guidelines:
    \begin{itemize}
        \item The answer NA means that the paper poses no such risks.
        \item Released models that have a high risk for misuse or dual-use should be released with necessary safeguards to allow for controlled use of the model, for example by requiring that users adhere to usage guidelines or restrictions to access the model or implementing safety filters. 
        \item Datasets that have been scraped from the Internet could pose safety risks. The authors should describe how they avoided releasing unsafe images.
        \item We recognize that providing effective safeguards is challenging, and many papers do not require this, but we encourage authors to take this into account and make a best faith effort.
    \end{itemize}

\item {\bf Licenses for existing assets}
    \item[] Question: Are the creators or original owners of assets (e.g., code, data, models), used in the paper, properly credited and are the license and terms of use explicitly mentioned and properly respected?
    \item[] Answer: \answerYes{} % Replace by \answerYes{}, \answerNo{}, or \answerNA{}.
    \item[] Justification: Proper citations are added to the paper. Licenses for used data is included in Appendix~\ref{app:licenses}.%\justificationTODO{}
    \item[] Guidelines:
    \begin{itemize}
        \item The answer NA means that the paper does not use existing assets.
        \item The authors should cite the original paper that produced the code package or dataset.
        \item The authors should state which version of the asset is used and, if possible, include a URL.
        \item The name of the license (e.g., CC-BY 4.0) should be included for each asset.
        \item For scraped data from a particular source (e.g., website), the copyright and terms of service of that source should be provided.
        \item If assets are released, the license, copyright information, and terms of use in the package should be provided. For popular datasets, \url{paperswithcode.com/datasets} has curated licenses for some datasets. Their licensing guide can help determine the license of a dataset.
        \item For existing datasets that are re-packaged, both the original license and the license of the derived asset (if it has changed) should be provided.
        \item If this information is not available online, the authors are encouraged to reach out to the asset's creators.
    \end{itemize}

\item {\bf New Assets}
    \item[] Question: Are new assets introduced in the paper well documented and is the documentation provided alongside the assets?
    \item[] Answer: \answerNA{} % Replace by \answerYes{}, \answerNo{}, or \answerNA{}.
    \item[] Justification: Our paper does not release new assets.%\justificationTODO{}
    \item[] Guidelines:
    \begin{itemize}
        \item The answer NA means that the paper does not release new assets.
        \item Researchers should communicate the details of the dataset/code/model as part of their submissions via structured templates. This includes details about training, license, limitations, etc. 
        \item The paper should discuss whether and how consent was obtained from people whose asset is used.
        \item At submission time, remember to anonymize your assets (if applicable). You can either create an anonymized URL or include an anonymized zip file.
    \end{itemize}

\item {\bf Crowdsourcing and Research with Human Subjects}
    \item[] Question: For crowdsourcing experiments and research with human subjects, does the paper include the full text of instructions given to participants and screenshots, if applicable, as well as details about compensation (if any)? 
    \item[] Answer: \answerNA{} % Replace by \answerYes{}, \answerNo{}, or \answerNA{}.
    \item[] Justification: Our research does not involve crowdsourcing or human subjects. %\justificationTODO{}
    \item[] Guidelines: 
    \begin{itemize}
        \item The answer NA means that the paper does not involve crowdsourcing nor research with human subjects.
        \item Including this information in the supplemental material is fine, but if the main contribution of the paper involves human subjects, then as much detail as possible should be included in the main paper. 
        \item According to the NeurIPS Code of Ethics, workers involved in data collection, curation, or other labor should be paid at least the minimum wage in the country of the data collector. 
    \end{itemize}

\item {\bf Institutional Review Board (IRB) Approvals or Equivalent for Research with Human Subjects}
    \item[] Question: Does the paper describe potential risks incurred by study participants, whether such risks were disclosed to the subjects, and whether Institutional Review Board (IRB) approvals (or an equivalent approval/review based on the requirements of your country or institution) were obtained?
    \item[] Answer: \answerNA{} % Replace by \answerYes{}, \answerNo{}, or \answerNA{}.
    \item[] Justification: Our research does not involve crowdsourcing or human subjects. %\justificationTODO{}
    \item[] Guidelines:
    \begin{itemize}
        \item The answer NA means that the paper does not involve crowdsourcing nor research with human subjects.
        \item Depending on the country in which research is conducted, IRB approval (or equivalent) may be required for any human subjects research. If you obtained IRB approval, you should clearly state this in the paper. 
        \item We recognize that the procedures for this may vary significantly between institutions and locations, and we expect authors to adhere to the NeurIPS Code of Ethics and the guidelines for their institution. 
        \item For initial submissions, do not include any information that would break anonymity (if applicable), such as the institution conducting the review.
    \end{itemize}

\end{enumerate}

\end{document}